\documentclass[lettersize,journal]{IEEEtran}
\usepackage{amsmath,amsfonts}
\usepackage{algorithmic}
\usepackage{algorithm}
\usepackage{array}
\usepackage[caption=false,font=normalsize,labelfont=sf,textfont=sf]{subfig}
\usepackage{textcomp}
\usepackage{stfloats}
\usepackage{url}
\usepackage{hyperref}
\usepackage{verbatim}
\usepackage{pifont}
\usepackage{graphicx}
\usepackage{cite}
\usepackage{multirow}
\usepackage{booktabs}
\usepackage{stfloats}
\usepackage{float}
\usepackage{afterpage}
\usepackage{listings}
\begin{document}
\bstctlcite{IEEEexample:BSTcontrol}

\title{OmniBridge: Unified Multimodal Understanding, Generation, and Retrieval via Latent Space Alignment}

\author{Teng Xiao, Zuchao Li, Lefei Zhang,~\IEEEmembership{Senior Member,~IEEE}
\thanks{Teng Xiao, Lefei Zhang are with the School of Computer Science, Wuhan University, Wuhan, China.
E-mail: {xiaoxiao, zhanglefei}@whu.edu.cn}
\thanks{Zuchao Li is the corresponding author at: School of Artificial Intelligence, Wuhan University, Wuhan, China. E-mail: zcli-charlie@whu.edu.cn}

}
\markboth{OmniBridge: Unified Multimodal Understanding, Generation, and Retrieval via Latent Space Alignment}%
{Teng \MakeLowercase{\textit{et al.}}: A Sample Article Using IEEEtran.cls for IEEE Journals}


\maketitle

\begin{abstract}
Recent advances in multimodal large language models (LLMs) have led to significant progress in understanding, generation, and retrieval tasks. However, current solutions often treat these tasks in isolation or require training LLMs from scratch, resulting in high computational costs and limited generalization across modalities. In this work, we present OmniBridge, a unified and modular multimodal framework that supports vision-language understanding, generation, and retrieval within a unified architecture. OmniBridge adopts a language-centric design that reuses pretrained LLMs and introduces a lightweight bidirectional latent alignment module. To address the challenge of task interference, we propose a two-stage decoupled training strategy: supervised fine-tuning and latent space alignment for aligning LLM behavior with multimodal reasoning, and semantic-guided diffusion training to align cross-modal latent spaces via learnable query embeddings. Extensive experiments across a wide range of benchmarks demonstrate that OmniBridge achieves competitive or state-of-the-art performance in all three tasks. Moreover, our results highlight the effectiveness of latent space alignment for unifying multimodal modeling under a shared representation space.
Code and models are released at \href{https://github.com/xiao-xt/OmniBridge}{https://github.com/xiao-xt/OmniBridge}.
\end{abstract}


\section{Introduction}

\IEEEPARstart{I}{n} recent years, the two fundamental pillars of multimodal intelligence—understanding and generation—have witnessed remarkable progress.
For multimodal understanding, large multimodal language models (MLLMs), such as LLaVA~\cite{llava}, have demonstrated impressive capabilities on vision-language tasks like visual question answering (VQA).
On the generation side, denoising diffusion probabilistic models (DDPMs)\cite{sohl2015deep,ho2020denoising} have revolutionized the traditional generation paradigm\cite{kingma2013auto,goodfellow2014generative}, achieving unprecedented performance in text-to-image and text-to-video synthesis~\cite{podell2023sdxl,esser2024scaling,ho2022video,wu2023tune}.
In parallel, multimodal retrieval—often tackled by specialized models—has also gained traction.
While traditional retrieval research primarily focused on unimodal or cross-modal scenarios, recent studies have increasingly emphasized multimodal retrieval, where both queries and candidates may involve joint image-text inputs.
This trend has driven widespread applications in instruction-based image search, document-level retrieval, and retrieval-augmented generation~\cite{wu2021fashion,liu2021image,MagicLens,chang2022webqa,DBLP:conf/iclr/LiuXL0023,luo2023end,yasunaga2023retrieval,VisRAG}.

\begin{figure}[h]
\centering
\includegraphics[width=\linewidth]{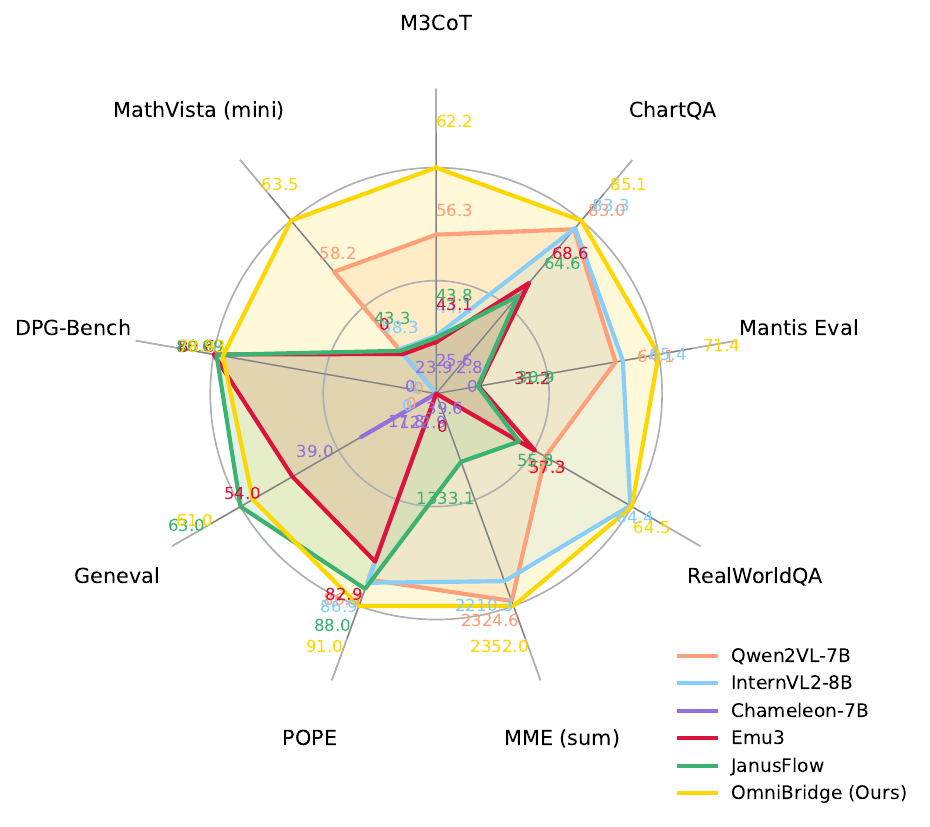}
\caption{\textbf{Benchmark Performances of multimodal understanding and image generation with OmniBridge}. OmniBridge surpasses the SOTA unified multimodal models and several task-specific understanding models on visual understanding benchmarks.}
\label{fig:radar}
\end{figure}

Given these advances in their respective subfields, a natural research direction is to explore unified frameworks that can bridge multimodal understanding and generation tasks.
Recent studies~\cite{NExT-GPT,SEED-X,X-VILA,tang2023any} have attempted to integrate expert models from different domains into a single system to handle both tasks concurrently.
However, many of these approaches still treat understanding and generation separately, often relying on distinct models for each.
For example, Chameleon~\cite{Chameleon} adopts a token-based autoregressive model trained on mixed image-text data, while TransFusion~\cite{Transfusion} and Show-o~\cite{Show-o} combine diffusion models with autoregressive methods to enhance generation quality.
Despite such efforts, these unified models often underperform compared to task-specific architectures, particularly in vision generation or complex reasoning tasks.
A notable advancement is Emu3~\cite{Emu3}, which demonstrates promising cross-modal generation via unified next-token prediction across images, videos, and texts.
Nevertheless, Emu3 requires up to 4096 tokens to generate a single image, resulting in significant computational overhead that limits its scalability.

Despite recent progress in unified multimodal modeling, two fundamental challenges remain unresolved, and they form the core motivation for this work:

\begin{itemize}
\item \textbf{Challenge 1: High Pretraining Cost and Computational Overhead.}
Models such as Chameleon, Show-o, and Emu3 have demonstrated the potential of unified architectures capable of handling both image and text modalities.
However, these models typically require training from scratch, involving extremely high computational costs to achieve balanced performance across modalities.
Even within a single modality, state-of-the-art (SOTA) models incur massive costs—for instance, pretraining a high-quality language model like LLaMA-3 demands over 15 trillion tokens.
Extending such training regimes to multimodal settings further amplifies the resource burden, hindering accessibility and reproducibility.

\item \textbf{Challenge 2: Difficulty in Balancing Understanding, Generation, and Retrieval.}  
While existing unified models perform well on either understanding or generation tasks, multimodal retrieval remains largely handled by specialized systems.  
Few models can simultaneously excel in all three modalities—understanding, generation, and retrieval—due to the inherent interference among tasks during joint training.  
In some cases, the objectives of these tasks may even conflict, leading to degraded overall performance.  
\end{itemize}

\begin{figure*}[ht]
\centering
\includegraphics[width=0.95\textwidth]{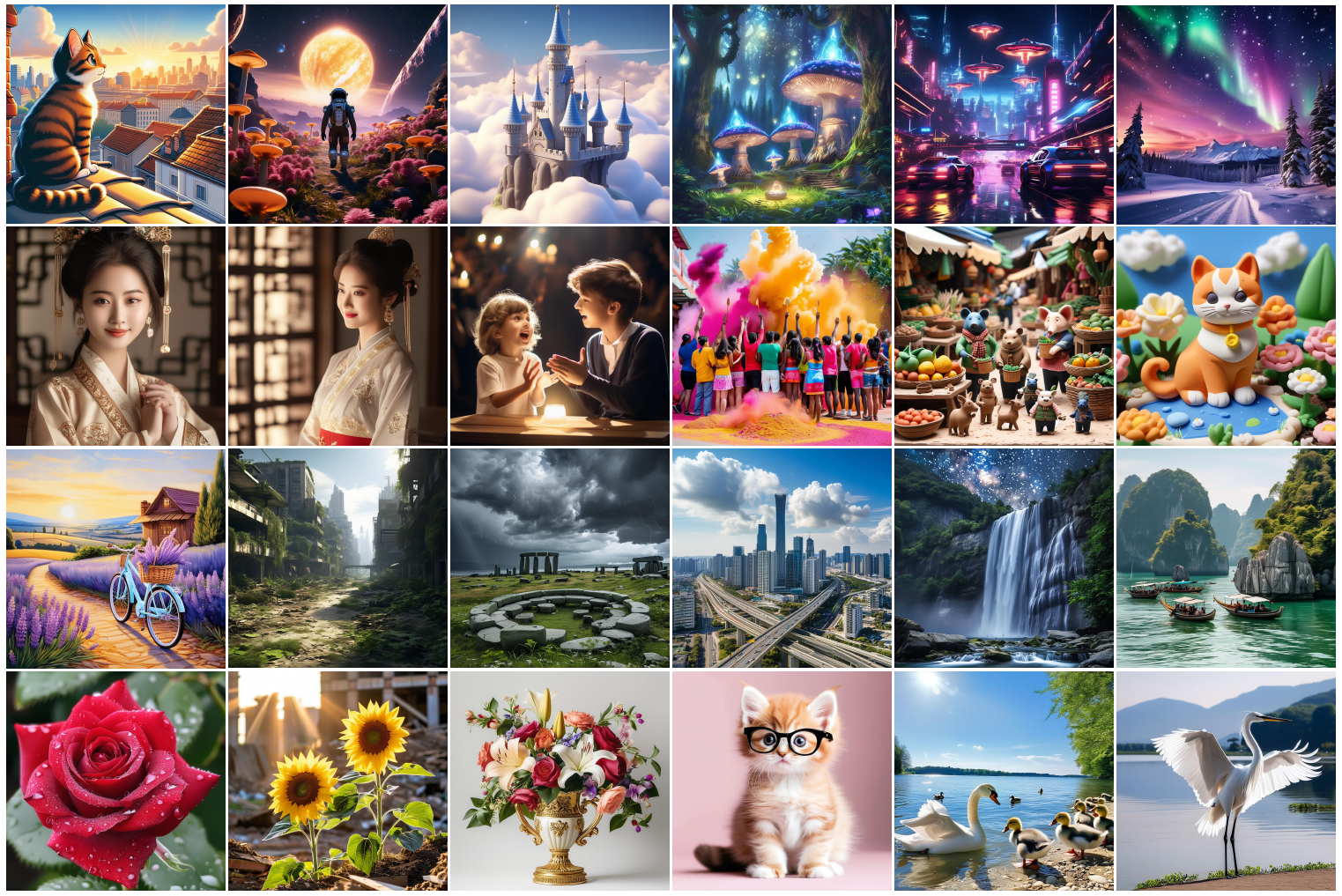}
\caption{Generated samples from OmniBridge, demonstrating strong visual generation capabilities.}
\label{fig:gen_examples}
\end{figure*}

To address these challenges, we propose \textbf{OmniBridge}, a unified and modular multimodal framework that combines a language-centric design with efficient cross-modal alignment.
Our approach avoids costly full-model pretraining by leveraging parameter-efficient adaptation techniques and reusing powerful pretrained LLMs.
With less than 100K training samples, OmniBridge can be trained effectively across multiple tasks using a decoupled training strategy that separates global reasoning from latent space alignment.
To tackle the first challenge of high computational cost, we introduce a lightweight adaptation mechanism based on LoRA, allowing the model to specialize in multimodal tasks without modifying the backbone LLM.
Instead of training from scratch, OmniBridge reuses a pretrained LLM and gradually aligns its behavior through high-quality supervised fine-tuning and reinforcement learning.
Furthermore, by decoupling visual generation, multimodal retrieval, and latent space alignment from the core LLM, our method enables efficient task adaptation while preserving the generalization capabilities of the language backbone.
To address the second challenge—balancing understanding, generation, and retrieval within a single model—we propose a hybrid architecture that combines a unidirectional LLM for autoregressive generation with a bidirectional Transformer (BiTransformer) for fine-grained latent alignment.
Unlike previous methods that treat these tasks separately, OmniBridge aligns both generation and retrieval within a shared latent space, ensuring semantic consistency and efficient feature matching across modalities.
The BiTransformer learns modality-invariant representations through a semantic-guided diffusion training strategy, enabling fine-grained alignment between visual and textual embeddings in the latent space.
This unified structure allows all three tasks to co-exist and mutually benefit from shared semantic priors, achieving strong performance without task conflict.

We evaluate OmniBridge extensively across a wide range of benchmarks covering multimodal understanding, generation, and retrieval tasks.
Experimental results provide strong evidence that latent space alignment—enabled by our bidirectional alignment module—is an effective mechanism for unifying multimodal learning in a single, coherent framework.
Notably, OmniBridge not only achieves competitive or SOTA performance on task-specific benchmarks, but also demonstrates consistent generalization across diverse modalities and objectives.
Specifically, Figure~\ref{fig:radar} shows that OmniBridge surpasses baseline models in multimodal understanding, while Figure~\ref{fig:gen_examples} showcases its ability to generate high-quality images.
Collectively, the main contributions of this paper are as follows:
\begin{itemize}
\item We propose OmniBridge, a unified and modular multimodal framework that supports understanding, generation, and retrieval tasks within a single architecture.

\item We introduce a two-stage decoupled training strategy that separates behavioral alignment from latent-level alignment, enabling efficient and stable adaptation across diverse multimodal tasks.

\item We design a novel semantic-guided diffusion training mechanism that gradually replaces text conditioning with learnable query embeddings, enabling fine-grained, controllable latent space alignment.

\item We demonstrate the effectiveness of our framework through extensive experiments on standard vision-language benchmarks, validating that OmniBridge has achieved state-of-the-art or competitive performance in multimodal understanding, generation, and retrieval tasks.
\end{itemize}

\section{Related Work}

\subsection{Multimodal Understanding}
Recent advances in Multimodal Large Language Models (MLLMs)~\cite{li2024multimodal,mllm2023Survey,qwen2vl} have significantly pushed the boundaries of visual and linguistic integration. Current architectures for multimodal understanding primarily follow two paradigms. One dominant approach connects a pre-trained visual encoder (e.g., CLIP~\cite{CLIP}) to an LLM, as seen in models like LLaVA~\cite{llava} and InstructBLIP~\cite{InstructBLIP}. While effective for complex reasoning, this paradigm is often bottlenecked by the high computational cost and slow inference of its large visual backbone. An alternative approach employs visual tokenizers~\cite{esser2021taming,Diffusion, Show-o} to convert images into discrete sequences, unifying the input format for both understanding and generation. However, this method typically involves complex multi-stage training, and the quantization process can lead to the loss of fine-grained visual details, complicating downstream alignment.

More recently, reinforcement learning (RL) techniques have been explored to further enhance the alignment and instruction-following capabilities of these models~\cite{dong2025insight, llavaReasoner}. Despite these advances, RL-based fine-tuning does not resolve the underlying architectural trade-offs. These methods still inherit the high computational costs of their base models and primarily focus on improving understanding-centric tasks.

\subsection{Vision Generation}


Autoregressive (AR) models, inspired by their success in NLP~\cite{transformer, llama}, have been adapted for visual synthesis by modeling sequences of visual tokens~\cite{esser2021taming, VideoPoet, LLamaGen}. However, this paradigm struggles with a trade-off between quality and efficiency. AR models often either lag behind diffusion models in generative fidelity or resort to complex, multi-stage architectures to remain competitive~\cite{VideoPoet}. Moreover, even state-of-the-art single-transformer models like Emu3~\cite{Emu3} suffer from severe computational inefficiency, requiring the processing of exceptionally long token sequences (e.g., 4096 tokens for one image), making them highly resource-intensive.

Diffusion models~\cite{SD, SDXL, PixArt-alpha} have emerged as the de facto standard for high-fidelity image synthesis. These models, typically implemented as U-Nets operating in a VAE's latent space, learn to denoise visual representations conditioned on multimodal inputs. Textual guidance is primarily injected via cross-attention layers~\cite{DALL-E2, Imagen}, a 'soft' alignment mechanism that also enables fine-grained spatial control in models like ControlNet~\cite{ControlNet} and PixArt-$\alpha$~\cite{PixArt-alpha}. Despite their success, this reliance on cross-attention poses a critical limitation: it often fails to enforce precise semantic correspondence, leading to errors in compositional reasoning and attribute binding.




\subsection{Multimodal Retrieval}

The scope of retrieval has expanded from traditional unimodal~\cite{BEIR} and cross-modal~\cite{MSCOCO} tasks to complex multimodal scenarios involving mixed-media queries and documents. This shift is driven by growing applications in retrieval-augmented generation~\cite{yasunaga2023retrieval, VisRAG} and instruction-based image retrieval~\cite{wu2021fashion, MagicLens}.

While many large models~\cite{qwen2vl,Emu3,li2023blip} exhibit multimodal capabilities, building a truly unified architecture that excels at understanding, generation, and retrieval remains a significant challenge. The core difficulty lies in reconciling conflicting objectives—the creative demands of generation versus the discriminative needs of retrieval—within a shared parameter space. This often leads to task interference during joint training. Thus, designing an architecture that enables harmonious coexistence and mutual enhancement of these diverse tasks is a key research direction.

\subsection{Unified Vision Language Models}
In recent years, an increasing number of studies~\cite{SEED-X,NExT-GPT, tang2023any, X-VILA} have focused on unified multimodal language models that can handle both understanding and generation tasks. 
One approach combines pre-trained diffusion models with multimodal large language models (LLMs)~\cite{DreamLLM, ge2023planting,SEED-X, Emu, X-VILA}. However, these systems essentially use diffusion models as external tools, where the LLM generates the necessary conditional text information for image generation but lacks direct generative capabilities. This separation often leads to performance degradation compared to standalone diffusion models \cite{ge2023planting, Emu}. Another research direction \cite{Chameleon, Janus, VILA-U, Show-o, Transfusion} aims to train a single LLM for both tasks. Many of these methods employ vector quantization~\cite{esser2021taming, sun2024autoregressive} to convert images into discrete tokens, enabling unified autoregressive processing. While these methods are simpler to implement, they require training additional visual tokenizers, which demand large amounts of data and computational power. Furthermore, due to limitations in image tokenization quality, the generation results from these methods are often inferior to those of diffusion-based models. Recent works, such as Chameleon~\cite{Chameleon}, have adopted token-based autoregressive models trained on mixed image and text data. TransFusion~\cite{Transfusion} and Show-o \cite{Show-o} attempt to combine diffusion models with autoregressive methods to enhance performance. However, these models still lag behind task-specific architectures in vision generation and understanding tasks. Emu3~\cite{Emu3} has achieved promising results for the first time through next-token prediction across images, video, and text, but it requires 4096 tokens to generate each image, which is highly resource-intensive. JanusFlow~\cite{JanusFlow} uses a flow-based method for image generation, which outperforms diffusion-based models in terms of image generation speed. However, in qualitative comparisons, it lags behind diffusion-based models in terms of the generated results.

In summary, most unified frameworks struggle with inherent task interference, failing to achieve a harmonious balance between understanding, generation, and retrieval. This landscape underscores the urgent need for a novel architectural paradigm that can reconcile these competing demands within a single, coherent, and efficient framework.

\section{Methodology}

\subsection{Overview}

We introduce OmniBridge, a unified multimodal autoregressive framework that supports a wide range of vision and vision-language tasks, including multimodal understanding, conditional image generation, and image-text retrieval.
Built on a language-centric architecture, OmniBridge integrates a LLM with a lightweight visual encoder and a bidirectional Transformer module (BiTransformer), enabling consistent and extensible cross-modal reasoning and generation.

\begin{figure*}[h]
\centering
\includegraphics[width=0.9\textwidth]{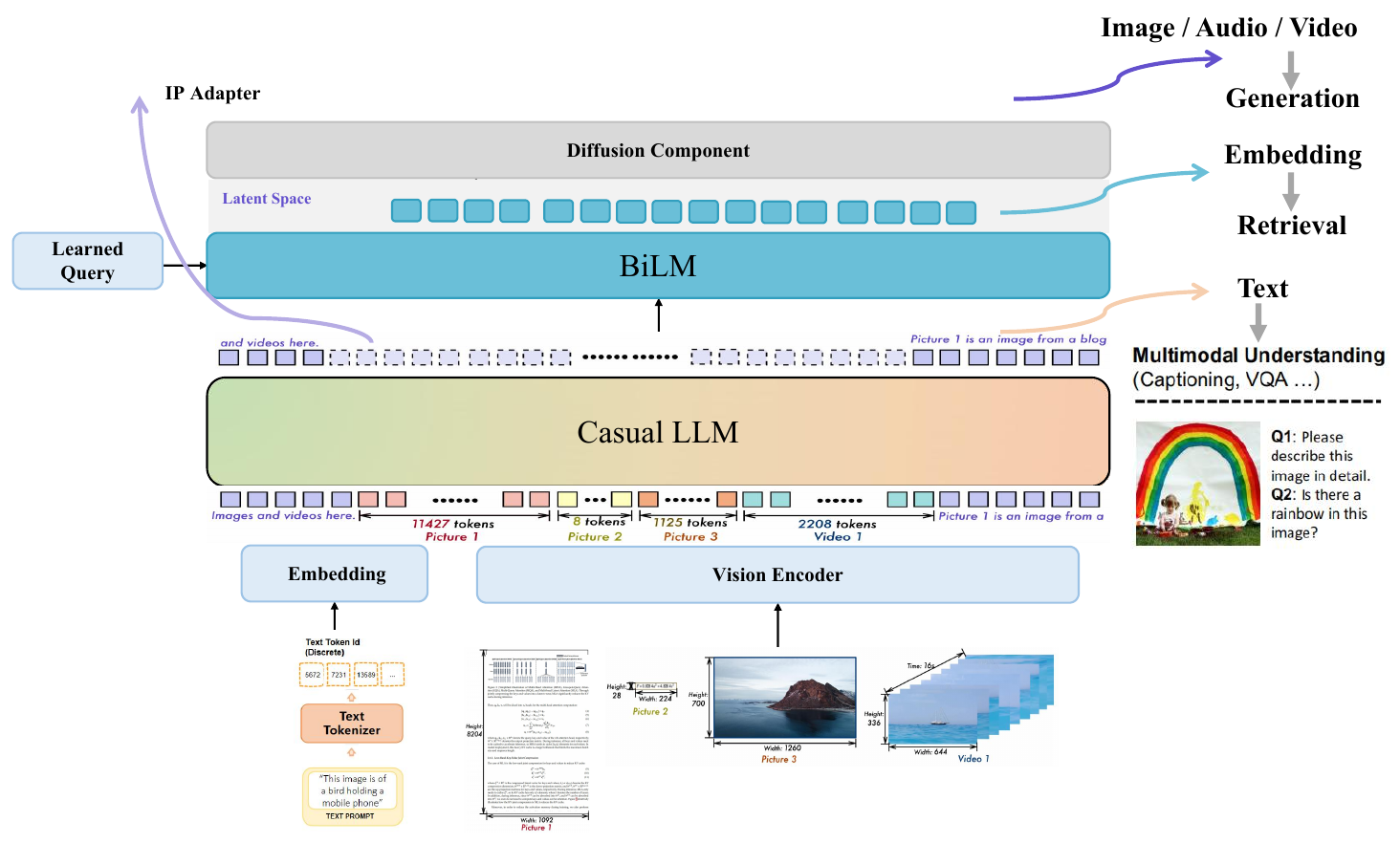}
\caption{\textbf{Overview of OmniBridge.} The framework consists of a unified language-centric backbone (LLM) supported by a visual encoder and a bidirectional Transformer module (BiTransformer). It supports three multimodal tasks—understanding, generation, and retrieval—through a two-stage decoupled training strategy: (1) supervised fine-tuning of the LLM for task-level alignment, and (2) latent space alignment using learnable queries and semantic-guided diffusion.}
\label{fig:framework}
\end{figure*}

As illustrated in Figure~\ref{fig:framework}, OmniBridge adopts a modular architecture that separates visual processing, language modeling, and latent alignment. Multimodal inputs—such as interleaved text and image features—are processed autoregressively by the LLM for reasoning tasks. For generative and retrieval tasks, a set of learnable generat queries and latent representations is introduced to guide visual outputs or align vision-language embeddings.

To support these diverse capabilities, we propose a two-stage decoupled training strategy:
\begin{itemize}
\item In \textit{Stage 1}, the LLM is adapted to multimodal instruction-following and reasoning tasks through high-quality supervised fine-tuning and reinforcement learning.
\item In \textit{Stage 2}, the latent alignment modules are trained with a semantic-guided diffusion strategy, gradually replacing explicit text supervision with internalized, query-based control for image generation, image editing, and multimodal retrieval tasks.
\end{itemize}

Together, these components allow OmniBridge to perform cross-modal tasks in a coherent and efficient manner, while preserving scalability and generalization across tasks.
A detailed description of each module and training phase is provided in the following sections.

\subsection{Unified Multimodal Autoregressive Modeling}

The core of OmniBridge is a unified, language-centric architecture that seamlessly integrates a bidirectional Transformer for cross-modal alignment. This design allows for versatile handling of a broad spectrum of multimodal tasks.

\textbf{\textit{For multimodal understanding tasks}}. OmniBridge processes an interleaved sequence of text and visual embeddings. Following standard practice, text inputs are tokenized and mapped to embeddings of dimension $D_{\mathrm{emb}}$. For each image $\mathbf{x}_{\mathrm{im}}$, a pre-trained visual encoder $f_{\mathrm{enc}}$ extracts a sequence of feature vectors. A lightweight projection layer then maps these visual features into the LLM's embedding space to ensure dimensional compatibility. The resulting unified sequence is fed into the LLM, which performs autoregressive prediction.






To improve the model’s reasoning ability in multimodal understanding, we first employ R1 distillation with long-form Chain-of-Thought (CoT) samples. This distillation process provides rich intermediate reasoning traces, encouraging the model to learn multi-step inference strategies. However, we observe that relying solely on R1 distillation can lead to repetitive or shallow reasoning in complex tasks. To mitigate this, we introduce a reinforcement learning phase inspired by policy optimization~\cite{zhang2025R1_vl}, which further refines the model’s output behavior and promotes diverse, accurate reasoning patterns.

\textbf{\textit{For generative tasks}}. The base LLM inherently exhibits strong image captioning capabilities. To extend its applicability to image generation and text-guided inpainting, we unify these tasks through a language-conditioning strategy. Specifically, to mitigate the limited capacity of diffusion models in processing long text prompts, we apply task-specific prompt rewriting rules. For image generation, short captions are expanded to emphasize spatial and relational attributes, while long, dense captions are compressed into concise and focused descriptions. For image editing, the input consists of an image and an editing instruction. The LLM generates a modified caption that reflects the desired post-edit image, which is then rewritten using the same prompt engineering rules.

The refined captions serve as the conditioning input for training the latent alignment module. Following common practice~\cite{Chameleon, Janus, Show-o}, we insert special tokens \texttt{<img>} and \texttt{</img>} around the caption to help the model identify and isolate guided embeddings within the sequence. This structured markup improves alignment between the textual condition and the visual output, especially in generative tasks that require spatial precision. Naturally, by including interleaved image-text samples in the training data, the model can generate paragraphs with interleaved visual and textual content. Since the underlying large language model uses text as the generative bridge, there is no explicit gap between modalities. The generated \texttt{image caption} can either be decoded as plain text or directly passed as hidden states into the bidirectional transformer to guide image generation.



\textbf{\textit{For multimodal retrieval tasks}}. Our goal is to achieve robust multimodal retrieval by aligning image and text representations within a shared embedding space. However, we depart from conventional dual-encoder architectures that rely on a dedicated visual encoder (e.g., a ViT) for images and a separate bidirectional text encoder (e.g., BERT) for text. Such approaches, while effective for retrieval, are inherently dis-unified and difficult to integrate with generative tasks.

In contrast, our approach performs alignment directly within the latent space of our unified model. Instead of encoding raw modalities from scratch, we leverage the rich, multimodal hidden states already produced by our base LLM. To achieve this, we introduce a powerful BiTransformer module specifically for fine-grained latent space alignment. This mechanism is realized through a novel \textbf{dual-path fusion architecture}, which combines deeply reasoned features from the BiTransformer with high-level semantic information from the base LLM. The entire process is optimized via a contrastive learning objective, enabling us to learn discriminative embeddings while preserving the model's unified structure for multi-task learning:

\begin{itemize}
\item \textbf{BiTransformer Path}. This primary path is designed to perform deep, contextualized reasoning over the base LLM's output features. We introduce a set of learnable, modality-specific queries ($\mathbf{Q}_{\text{img}} \in \mathbb{R}^{N_q^{img} \times D}$ and $\mathbf{Q}_{\text{text}} \in \mathbb{R}^{N_q^{text} \times D}$). These queries are processed by a BiTransformer module, which takes the final layer hidden states from the base LLM ($\mathbf{H}_{\text{img}}$ and $\mathbf{H}_{\text{text}}$) as key and value pairs for its cross-attention mechanism. The bidirectional nature of this module is crucial: it allows each query token to attend to the \textit{entire sequence} of LLM hidden states simultaneously. This enables a holistic aggregation and refinement of information, producing highly contextualized query embeddings. These embeddings are then passed through an attention pooling layer to yield the final BiTransformer path representations, $\mathbf{e}_{\text{img}}^{\text{BiT}}$ and $\mathbf{e}_{\text{text}}^{\text{BiT}}$.

\item \textbf{Direct Backbone Path}. Complementing the deep reasoning path, this second path provides a direct, global representation from the base model. We apply a separate attention pooling layer directly to the LLM's hidden states ($\mathbf{H}_{\text{img}}$ and $\mathbf{H}_{\text{text}}$) to obtain the backbone embeddings, $\mathbf{e}_{\text{img}}^{\text{LLM}}$ and $\mathbf{e}_{\text{text}}^{\text{LLM}}$. This path ensures that the rich, high-level semantics captured by the powerful base LLM are preserved.

\item \textbf{Learnable Fusion and Objective.} To dynamically leverage the strengths of both paths, we fuse their outputs via a learnable, weighted average:
\begin{equation}
\begin{aligned}
\mathbf{e}_{m}
&= \sigma(\alpha_{m})\,\mathbf{e}_{m}^{\mathrm{BiT}}
 + \bigl(1-\sigma(\alpha_{m})\bigr)\,\mathbf{e}_{m}^{\mathrm{LLM}},\\
&\text{where } m\in\{\mathrm{img},\mathrm{text}\}.
\end{aligned}
\end{equation}
where $\alpha_{\text{img}}$ and $\alpha_{\text{text}}$ are learnable scalar parameters and $\sigma$ is the sigmoid function to ensure the weights are between 0 and 1. The final fused embeddings, $\mathbf{e}_{\text{img}}$ and $\mathbf{e}_{\text{text}}$, are L2-normalized and optimized using a symmetric InfoNCE contrastive loss ($\mathcal{L}_{\text{ITC}}$).

\end{itemize}

\subsection{Latent Space Alignment via Bidirectional Transformer}

To enable controllable latent space alignment in generation tasks, we introduce a two-part strategy consisting of 1) learned query embeddings and 2) a semantic-guided diffusion training scheme.

\subsubsection{Learned Query Embeddings}

We employ a set of learnable query vectors as input to the BiTransformer.  These queries interact with the LLM's hidden states via cross-attention and with each other via self-attention, enabling the extraction of structured, task-specific features. Individual queries can thus specialize in capturing generation-relevant signals such as object layout, visual style, or editing intent.

\subsubsection{Semantic-Guided Diffusion Training}

To shift the model's conditional dependence from explicit text to these latent queries, we propose a Semantic-Guided Diffusion Training strategy.  The core idea is to progressively replace the text conditioning with our learnable queries throughout training. This process compels the model to internalize semantic control, conceptually resembling a diffusion process where the reliance on external text supervision is gradually "denoised."

\textbf{Progressive Replacement Schedule}. This is implemented via a three-stage schedule governed by a mixing coefficient $\beta$, as shown in Eq.~\ref{eq:replacement}:

\begin{itemize}
\item{
Initial Stage (15\% replacement): The model is primarily conditioned on text to ensure strong semantic grounding while introducing the query mechanism.
}
\item{
Progressive Stage (15\% → 75\%): The replacement ratio is linearly increased, forcing a gradual shift toward latent control without catastrophic forgetting.
}
\item{
Final Stage (100\% replacement): The model operates purely on the learned query embeddings for generation.
}
\end{itemize}

\begin{align}
\mathbf{z}_{\mathrm{cond}}=\beta\cdot\mathbf{z}_{\mathrm{text}}+(1-\beta)\cdot\mathbf{z}_{\mathrm{query}}
\label{eq:replacement}
\end{align}
Here, $\mathbf{z}_{\mathrm{text}}$ and $\mathbf{z}_{\mathrm{query}}$ are the text and learnable query embeddings, respectively. The mixing coefficient $\beta$ is annealed from a high value towards 0 during training.

This gradual substitution strategy ensures the model effectively transfers semantic priors from the text into the learnable queries. The resulting generation module is thus robustly controllable via its latent space, maintaining high fidelity even with weak or absent textual prompts at inference.


\subsection{Two-Stage Decoupled Training}

\begin{figure*}[h]
\centering
\includegraphics[width=0.9\textwidth]{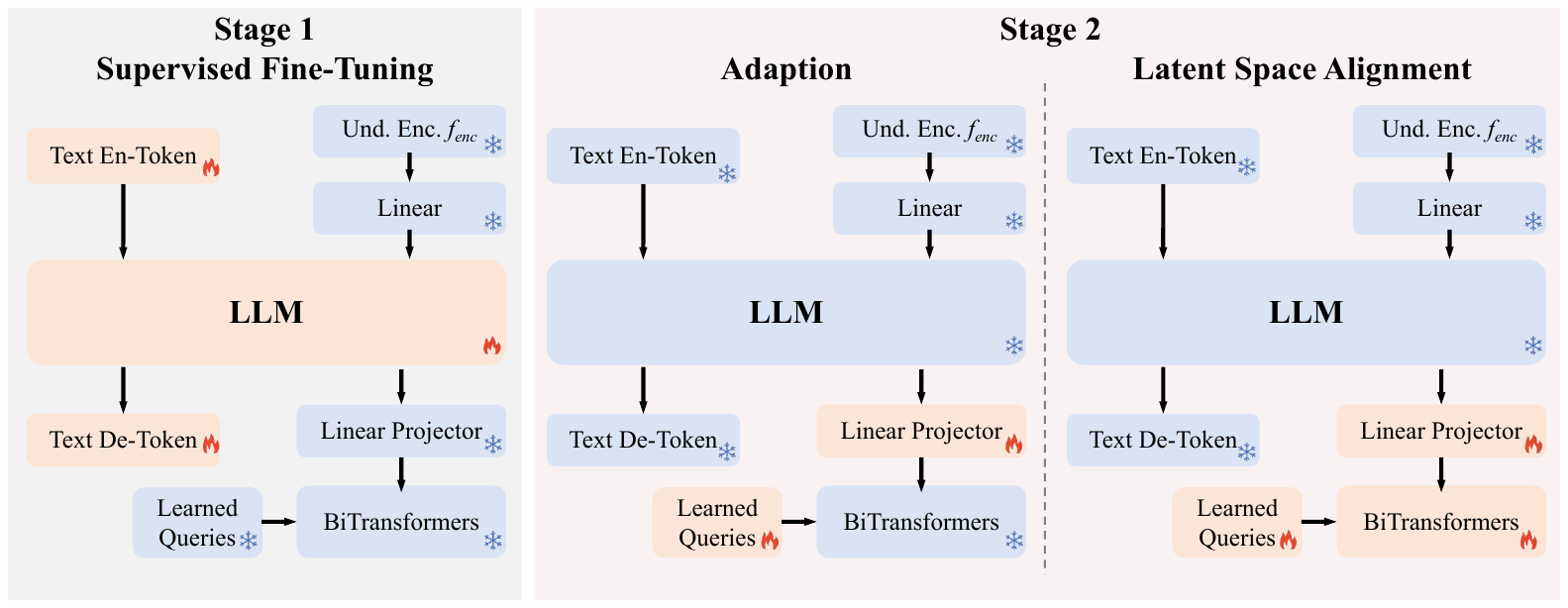}
\caption{\textbf{Two training stages of OmniBridge}. The trainable modules are marked with flame
and the frozen modules are marked with snowflakes.}
\label{fig:stage}
\end{figure*}

To mitigate the task interference inherent in unified modeling, we propose a two-stage decoupled training strategy, illustrated in Figure~\ref{fig:stage}. This approach first adapts the LLM for high-level reasoning and instruction following, then separately optimizes the latent space for fine-grained generation and retrieval. By isolating global behavioral alignment from modality-specific representation learning, our strategy ensures stable convergence and enhances multi-task performance.


\subsubsection{Stage 1: High-Quality Supervised Fine-tuning} 

In the first stage, we align the LLM to high-level multimodal understanding using LoRA-based R1 distillation and StepGRPO~\cite{zhang2025R1_vl}. R1 transfers fine-grained reasoning traces (e.g., long-form Chain-of-Thought) so the model internalizes structured strategies, but—since R1 distillation alone can yield repetitive or suboptimal behavior on complex cases—we further apply StepGRPO to refine the output policy toward concise, accurate, task-aligned responses. Training uses high-quality vision–language data spanning image description, scene editing, and instruction following, while the latent alignment module is frozen, allowing the LLM to focus on task-level objectives under strong supervision and thereby enhancing both reasoning ability and semantic alignment.

\subsubsection{Stage 2: Latent Space Alignment}

In the second stage, we learn a structured latent representation for generation and retrieval while keeping the LLM frozen.  We first adapt the randomly initialized modules—linear projection layers, learnable query embeddings, and cross-attention—so they interoperate with the pretrained LLM and the BiTransformer; this initialization stabilizes training before full alignment. We then jointly train the BiTransformer and these modules using the previously described Semantic-Guided Diffusion Training, gradually shifting conditioning from explicit text embeddings to learned latent queries and thereby encoding task-specific semantics in a structured, controllable latent space.

\subsection{Training Objective}

The training of OmniBridge is guided by a unified objective comprising three specialized losses, each mapped to a specific stage of our decoupled training strategy to mitigate task interference and ensure stable convergence.

\textbf{Autoregressive Loss for Multimodal Understanding}. For high-level understanding and reasoning tasks, the LLM is optimized with a standard autoregressive loss ($\mathcal{L}_{\mathrm{AR}}$) over the textual output:


\begin{align}
\mathcal{L}_{\mathrm{AR}} = - \mathbb{E}_{x \sim \mathcal{D}_{\mathrm{und}}} \left[ \sum_{i = \ell_{\mathrm{con}}}^{\ell - 1} \log P_{\theta}(x_{i+1} \mid x_1, \ldots, x_i) \right]
\label{eq:ar_loss}
\end{align}
where $\ell_{\mathrm{con}}$ is the length of the input context (image and instruction). This loss is exclusively used in Stage 1 to align the LLM's behavior with task-specific objectives like visual question answering.


\textbf{Latent Regression Loss for Text-Conditional Image Generation}. To endow the model with generation capabilities, we supervise the prediction of visual latent variables in a diffusion-like manner. Given the condition input (query or text embedding) and the target latent $\mathrm{z}_t$, we minimize:
\begin{align}
\mathcal{L}_{\mathrm{Gen}} = \mathbb{E}_{\mathbf{z}_0 \sim q,\, t \sim \mathcal{U}(0,1)} \left[ \left\| \mathbf{z}_t - \hat{\mathbf{z}}_t(\mathbf{c}) \right\|_2^2 \right]
\label{eq:mse_loss}
\end{align}
Here, $\hat{\mathbf{z}}_t(\mathbf{c})$ is the predicted latent from the BiTransformer conditioned on $\mathbf{c}$, which gradually transitions from text embedding to learned query embedding under the semantic-guided diffusion strategy in Stage 2.

\textbf{Contrastive Loss for Multimodal Retrieval}. Multimodal retrieval is optimized via an InfoNCE-style contrastive loss ($\mathcal{L}_{\mathrm{ITC}}$). Given image embedding $z_{\mathrm{img}}$ and text representation $z_{\mathrm{text}}$, the loss is defined as:
\begin{align}
\mathcal{L}_{\mathrm{ITC}} = - \log \frac{\exp(\mathrm{sim}(z_{\mathrm{img}}, z_{\mathrm{text}})/\tau)}{\sum_{j=1}^{N} \exp(\mathrm{sim}(z_{\mathrm{img}}, z_{\mathrm{text}}^{j})/\tau)}
\label{eq:ltc_loss}
\end{align}
where $sim(,)$ denotes cosine similarity, $\tau$ is a temperature hyperparameter, and $z_{text}$ is the positive sample among $N$ in-batch candidates. The image embedding and text embedding are obtained via the BiTransformer.

\textbf{Overall Objective}. The total training objective is a composition of these losses, applied selectively according to our decoupled strategy. Stage 1 exclusively uses $\mathcal{L}_{\mathrm{AR}}$ to adapt the LLM's behavior. In Stage 2, with the LLM frozen, $\mathcal{L}_{\mathrm{Gen}}$ and $\mathcal{L}_{\mathrm{ITC}}$ are used to train the latent space modules for their respective tasks. Detailed experimental settings are provided in Sec.~\ref{sec:exp:setup}.


\section{Experiments}
We conduct extensive and systematic experiments to comprehensively evaluate the performance of OmniBridge across multimodal understanding, generation, and retrieval tasks. We begin by detailing the experimental setup and implementation. Next, we present results on standard benchmarks for multimodal understanding, image generation, and retrieval. Finally, ablation studies are performed to assess the contributions of key design components.

\subsection{Experiment Setup and Implementation Details}
\label{sec:exp:setup}

Our framework is built upon the Qwen2-VL-7B-Instruct model~\cite{qwen2vl}, a language model composed of 28 transformer blocks with support for a maximum sequence length of 32,768 tokens. Benefiting from Qwen2-VL’s ability to handle images of arbitrary resolution, our model imposes no restrictions on image resolution for both understanding and generation tasks.

To enable multimodal understanding, we inherit the vision-language capabilities of Qwen2-VL. For image generation, we adopt a pretrained mT5 model~\cite{mt5} (approximately 700M parameters) as the bidirectional backbone to facilitate latent space alignment. A linear downsampling layer is used to project the hidden states from Qwen2-VL into the input space of the bidirectional model. Cross-attention modules are inserted at layers (i.e., 0, 2, 4, 8, 12, 16, and 20) of the bidirectional transformer to learn joint multimodal representations. To enable unconditional image generation, we introduce a set of learnable queries that guide the model without textual input to BiTransformers. Finally, the HunyuanDiT decoder~\cite{HunyuanDiT} is employed to complete the diffusion process and synthesize the output image.

For data preprocessing, no further adjustment is required once high-quality samples are obtained. The first stage of training uses a mixture of tasks including multimodal understanding, image generation, and image editing. In the second stage, only generation, editing, and multimodal retrieval data are used, with understanding tasks excluded.
Our implementation is based on the Linux platform and PyTorch~\cite{pytorch2024}, and the training is conducted using the ms-swift. All models are trained on NVIDIA A800 GPUs, with each model requiring fewer than 10 A800 GPU days (without reinforcement learning).

\subsection{Training Data Settings}
\label{sec:exp:data}
We generate and adopt a small amount of open-source data to construct the training data. The data configuration for each training stage is listed below.

Data for Stage 1 and Stage 2. The first two stages of our framework use five types of data:
multimodal understanding data, unimodal understanding data, image generation data, image editing data, and multimodal retrieval data.

\begin{itemize}
  \item Multimodal understanding data. We randomly sample 7000 data from Mulberry-260k~\cite{yao2024mulberry} as multimodal understanding. Furthermore, Mulberry-260k is selected for  reinforcement learning training.
  \item Unimodal understanding data. We select 2000 data from Magpie-Reasoning-V2-250K-CoT-Deepseek-R1\footnote{\href{https://huggingface.co/datasets/Magpie-Align/Magpie-Reasoning-V2-250K-CoT-Deepseek-R1-Llama-70B}{Magpie-Reasoning-V2-250K-CoT-Deepseek-R1-Llama-70B dataset}} by sequence length for unimodal understanding ability.
  \item Image generation data. To enable the model to handle both short and long text prompts for image generation, we used Qwen3-30B-A3B~\cite{yang2025qwen3} to generate 3,000 short captions and 4,000 long captions across a wide range of categories for Stage 1 and to generate 19,000 short captions and 19,000 long captions for Stage 2.
  \item Image editing data. 2500 samples and 5500 samples in MagicBrush~\cite{zhang2023magicbrush} are selected for for Stage 1 and Stage 2, respectively.
  \item Multimodal retrieval data. We directly incorporate the multimodal retrieval data from Flickr30K~\cite{plummer2015flickr30k}.
\end{itemize}







\subsection{Evaluation Details}
\label{sec:exp:eval}

\textbf{Multimodal Understanding.} 
To comprehensively evaluate OmniBridge's performance on multimodal understanding tasks, we employ a diverse set of benchmarks from both established and recent collections, including VLMEvalKit~\cite{Vlmevalkit}, spanning diverse task categories:

\begin{itemize}
  \item Multimodal Reasoning and Mathematics.
  We comprehensively assess OmniBridge’s multimodal math and reasoning across domain benchmarks: MMMU~\cite{MMMU}, MMMU-Pro~\cite{MMMUPro}, ScienceQA~\cite{ScienceQA}, MathVista~\cite{MathVista}, MATH-Vision~\cite{MATHVision}, MathVerse~\cite{MathVerse}, and M3CoT~\cite{M3CoT}. For MMMU-Pro, we report three metrics: standard (10 options), vision, and overall (the average of the two). 
  
  
  \item OCR, Chart, and Document Understanding. We assess OmniBridge's capabilities on a wide array of OCR-centric tasks: AI2D~\cite{AI2D} (mask/no-mask settings), ChartQA~\cite{ChartQA} (test accuracy), CharXiv~\cite{Charxiv} (RQ/DQ tasks), TextVQA~\cite{TextVQA} (validation VQA accuracy), DocVQA~\cite{DocVQA}, InfoVQA~\cite{InfoVQA}, OCRBench~\cite{OCRBench}, and SEEDBench-2-Plus~\cite{SEEDBench-2-Plus}. Performance on DocVQA and InfoVQA is measured using the ANLS metric.
  
  \item Multi-Image Understanding. We evaluate OmniBridge’s multi-image relational reasoning on a suite of benchmarks: BLINK~\cite{BLINK}, Mantis-Eval~\cite{MANTIS}, MuirBench~\cite{MuirBench}, and MMT-Bench~\cite{MMT-Bench}. For BLINK and MMT-Bench, we report results on their respective validation sets.
  
  \item Real-World Comprehension. We assess OmniBridge on real-world suites targeting complex, realistic tasks: RealWorldQA~\cite{RealWorldQA}; MME-RealWorld~\cite{MME-RealWorld}, for which we use the English full set; and R-Bench~\cite{R-Bench}, reporting the score for the ``R-Bench-Dis'' setting in VLMEvalKit.


  \item Comprehensive Multimodal Evaluation. OmniBridge's comprehensive capabilities are evaluated on several standard benchmarks: MME~\cite{MME}, for which we report the overall score; MMBench~\cite{Mmbench}, using the English version's test set; and the recently proposed MMStar~\cite{MMStar}.
  
  \item Multimodal Hallucination Evaluation. To evaluate the model's tendency toward hallucination, we employ two standard benchmarks: HallusionBench~\cite{HallusionBench}, reporting the average of its aAcc, fAcc, and qAcc scores; and POPE~\cite{POPE}, reporting the average F1 score.
\end{itemize}

\textbf{Multimodal Generation.} We quantify text-to-image performance via semantic accuracy—fidelity to prompt-specified objects and relations—and benchmark OmniBridge on GenEval~\cite{Geneval} and DPG-Bench~\cite{DPG-Bench}.


\textbf{Image Editing.} To evaluate the image editing capability, we employ ImgEdit~\cite{ye2025imgedit}, which recently introduced benchmarks featuring 811 image-instruction
pairs.

\subsection{Multimodal Understanding}



\subsubsection{Multimodal Reasoning and Mathematics}

\textbf{Multidisciplinary Reasoning.} The left section of Table \ref{tab:exp:reasoning-benchmarks} benchmarks OmniBridge against three distinct MLLM categories. The results show consistent and significant improvements across the board.
First, OmniBridge achieves a +7.8\% absolute gain on ScienceQA-Img over its Qwen2-VL-7B backbone, demonstrating the effectiveness of our training strategy.
Second, it surpasses specialized CoT models like LLaVA-CoT-11B~\cite{xu2024llavacot} and Mulberry-7B~\cite{yao2024mulberry}, a success we attribute to our use of R1 distillation, which substantially strengthens complex reasoning.
Third, OmniBridge sets a new state-of-the-art among unified models, outperforming prior leaders like JanusFlow~\cite{JanusFlow} and Emu3~\cite{Emu3}.
This result is crucial, as it proves that our approach adds generative capabilities without compromising—and in fact, significantly enhancing—the base model's core understanding abilities.


\textbf{Mathematical Reasoning.} The right-hand section of Table \ref{tab:exp:reasoning-benchmarks} highlights OmniBridge's strong mathematical reasoning, a domain where unified models typically falter.
OmniBridge achieves 63.5\% on the MathVista test-mini and surpasses most specialized understanding and reasoning baselines on benchmarks like MATH-Vision and MathVerse. 
We attribute this distinct advantage primarily to our RL phase. While R1 distillation alone successfully instills CoT reasoning, we observed that it can lead to repetitive or erroneous steps in complex mathematical problems. The subsequent RL fine-tuning directly mitigates this failure mode by refining the model's policy, enabling it to produce more accurate and concise final answers.

\begin{table*}[t]
\centering
\caption{\textbf{Comparison of multimodal reasoning and mathematical performance}. MMMU~\cite{MMMU},
MMMU-Pro~\cite{MMMUPro}, and ScienceQA-Img~\cite{ScienceQA}  are multidisciplinary reasoning benchmarks, while MathVista~\cite{MathVista}, MATH-Vision~\cite{MATHVision}, and 
MathVerse~\cite{MathVerse} are mathematics benchmarks. Part of results are collected from~\cite{deitke2024molmo_54, MMMUPro,wang2024measuring_245,MathVerse, internvl2.5}
and the OpenCompass leaderboard~\cite{Opencompass}.}
\resizebox{0.85\linewidth}{!}{
\begin{tabular}{l|cccc|ccccc}
\toprule
\textbf{Model} &
\begin{tabular}{@{}c@{}}MMMU\\(val)\end{tabular} &
\begin{tabular}{@{}c@{}}MMMU\\(test)\end{tabular} &
\begin{tabular}{@{}c@{}}MMMU-Pro\\(std10 / vision / overall)\end{tabular} &
\begin{tabular}{@{}c@{}}ScienceQA\\-Img\end{tabular} &
\begin{tabular}{@{}c@{}}MathVista\\(mini)\end{tabular} &
\begin{tabular}{@{}c@{}}MATH-Vision\\(mini / full)\end{tabular} &
\begin{tabular}{@{}c@{}}MathVerse\\(mini)\end{tabular} \\

\midrule
\multicolumn{8}{l}{\textit{Open-Source Model (Understanding)}} \\
Phi-3.5-Vision-4B~\cite{abdin2024phi3technicalreporthighly} & 43.0 & - & 26.3 / 13.1 / 19.7 & - & 43.9 & 17.4 / 15.5 & 24.1 \\
Ovis1.6-Gemma2-9B~\cite{lu2024ovis} & 55.0 & - & - & - & 67.2 & - / 18.8 & - \\
MiniCPM-V2.6~\cite{yao2024minicpm} & 49.8 & - & 30.2 / 24.2 / 27.2 & - & 60.6 & 16.1 / 17.5 & 25.7 \\
InternVL2-8B~\cite{internvl2} & 52.6 & 44.3 & 32.5 / 25.4 / 29.0 & 97.2 & 58.3 & 20.4 / 18.4 & 37.0 \\
InternVL2.5-8B~\cite{internvl2.5} & 56.0 & 48.9 & 38.2 / 30.4 / 34.3 & - & 64.4 & 22.0 / 19.7 & 39.5 \\
Qwen2-VL-7B~\cite{qwen2vl} & 53.7 & 46.8 & 34.1 / 27.0 / 30.5 & 85.5 & 58.2 & 23.4 / 19.5 & 31.9 \\

\midrule
\multicolumn{8}{l}{\textit{Reasoning Model}} \\
Mulberry-7B~\cite{yao2024mulberry}  & 55.0 & - & - & 63.1 & - & - & - \\
LLaVA-CoT-11B~\cite{xu2024llavacot} & - & - & - & - & 54.8 & - & -   \\
LLaVA-Reasoner-8B~\cite{llavaReasoner} & 40.0 & - & - & 52.1 & - & -  \\
Insight-V-8B~\cite{dong2025insight} & 50.2 & - & 24.9 / - / - & - & 59.9 & -  & -   \\
LlamaV-o1-11B~\cite{thawakar2025llamavo1} & - & - & - & - & 54.4 & - & - \\

\midrule
\multicolumn{8}{l}{\textit{Open-Source Model (Unified)}} \\
Chameleon-7B~\cite{Chameleon} & - & - & - & 56.0 & 23.9 & 15.5 /  & 5.6  \\ 
Show-o~\cite{Show-o} & 25.1 & 27.4 & - & - & - & - & -  \\
EVE-7B (HD)~\cite{diao2024unveiling} & - & - & - & 62.6 & - & - & -  \\
Emu3~\cite{Emu3} & 32.3 & 29.0 & - & 87.9 & 47.3 & 17.4 / 15.5 & -  \\
JanusFlow~\cite{JanusFlow} & 32.9 & 30.1 & 17.9 / - / - & 60.3 &  43.3 & 18.1 / 14.5 & 11.5  \\
\textbf{OmniBridge (Ours)} & 52.3 & 47.2 & 36.2 / 27.5 / 31.9 & 93.3 & 63.5 & 25.0 / 24.9 & 38.5  \\

\bottomrule
\end{tabular}
}
\label{tab:exp:reasoning-benchmarks}
\end{table*}



On the challenging M3CoT benchmark for complex multimodal reasoning, OmniBridge demonstrates superior performance, as detailed in Table~\ref{tab:m3CoT}. It achieves state-of-the-art results across multiple domains, including Science and Commonsense, with particularly strong scores in language-based science (86.3\%) and physics-based reasoning (90.0\%). With an overall accuracy of 62.2\%, OmniBridge significantly outperforms all baselines, including specialized reasoning models like LLaVA-CoT-11B (55.6\%) and its own base model, Qwen2VL-7B (56.3\%).

\begin{table}[t]
\centering
\caption{Comparison of Multimodal, multi-hop and multi-domain reasoning performance on M3CoT~\cite{M3CoT} Benchmark.}
\resizebox{\linewidth}{!}{
\begin{tabular}{lcccccccccc}
\toprule
\multirow{2}{*}{\textbf{Model}} & 
\multicolumn{3}{c}{\textbf{Science}} & 
\multicolumn{3}{c}{\textbf{Commonsense}} & 
\multicolumn{3}{c}{\textbf{Mathematics}} & 
\multirow{2}{*}{\textbf{Total}} \\
\cmidrule(lr){2-4} \cmidrule(lr){5-7} \cmidrule(lr){8-10}
& Lang & Nat & Soc & Phy & Soc & Temp & Alg & Geo & Theo & \multicolumn{1}{c}{} \\
\midrule

\multicolumn{11}{l}{\textit{Open-Source Model (Understanding)}} \\
InternVL2-8B~\cite{internvl2} & 46.5 & 47.6 & 30.9 & 76.7 & 65.7 & 72.4 & 12.9 & 22.5 & 14.3  & 44.1 \\
Qwen2VL-7B~\cite{qwen2vl} & 72.9 & 56.3 & 43.3 & 77.8 & 79.6 & 78.7 & 47.9 & 31.3 & 56.7 & 56.3 \\
InstructBLIP-13B~\cite{InstructBLIP} & 38.4 & 30.0 & 27.6 & 80.0 & 70.3 & 33.3 & 30.7 & 21.3 & 19.1 & 36.1 \\
CogVLM-17B~\cite{wang2024cogvlm} & 51.2 & 43.8 & 29.3 & 54.4 & 39.3 & 31.7 & 35.7 & 33.8 & 33.3  & 38.9 \\
InternVL2-26B~\cite{internvl2} & 69.7 & 49.7 & 33.0 & 80.0 & 64.9 & 73.2 & 15.0 & 17.5 & 19.1 & 47.5 \\

\midrule
\multicolumn{11}{l}{\textit{Reasoning Model}} \\
MM-CoT~\cite{mmcot}   & 44.1 & 30.9 & 29.1 & 34.4 & 19.8 & 23.6 & 27.9 & 35.0 & 42.9 & 30.3 \\
Visual-CoT~\cite{shao2024visual}  & 42.2 & 31.4 & 24.0 & 73.3 & 65.3 & 41.5 & 24.3 & 41.3 & 4.8  & 35.8 \\
LlamaV-o1-11B~\cite{thawakar2025llamavo1}   & 16.1 & 53.0 & 35.4 & 75.6 & 73.1 & 67.5 & 30.0 & 17.5 & 14.3 & 45.6 \\
LLaVA-CoT-11B~\cite{xu2024llavacot} & 71.6 & 57.2 & 37.6 & 84.4 & 74.5 & 80.5 & 45.0 & 31.3 & 47.6 & 55.6 \\

\midrule
\multicolumn{11}{l}{\textit{Open-Source Model (Unified)}} \\

Chameleon-7B~\cite{Chameleon} &  11.4 &  23.6 & 24.2 & 44.4 & 54.5 & 26.8 & 9.3 & 13.8 & 19.0 & 25.6 \\
Emu3~\cite{Emu3} & 43.6 & 42.8 & 30.1 & 87.8 & 61.9 & 68.3 & 0.3 & 0.21 & 0.52 & 43.1\\
JanusFlow~\cite{JanusFlow} & 44.1 & 43.7 & 36.8 & 82.2 & 66.5 & 61.4 & 31.4 & 33.7 & 56.4 &  43.8 \\ 
\textbf{OmniBridge (Ours)} & 86.3 & 63.6 & 46.0 & 90.0 & 78.5 & 89.4 & 44.3 & 27.5 & 38.1 & 62.2 \\

\bottomrule
\end{tabular}
}
\label{tab:m3CoT}
\end{table}

This strong cross-domain generalization validates our training strategy. We attribute this success in complex reasoning primarily to our RL phase. Specifically, while R1 distillation alone is effective at instilling CoT patterns, it can lead to repetitive reasoning on complex problems. The subsequent RL fine-tuning directly mitigates this failure mode by rewarding concise and accurate paths to the final answer, enhancing the model's inference-time efficiency and correctness.

\subsubsection{OCR, Chart, and Document Understanding}

As shown in Table~\ref{tab:vqa-docqa-benchmarks}, OmniBridge exhibits a nuanced performance profile on these tasks. The model demonstrates clear strengths in structured reasoning, improving upon its Qwen2VL-7B base by +2.1\% on ChartQA. This gain highlights the success of our targeted enhancements for symbolic alignment and numerical reasoning. Conversely, we observe a modest performance drop on OCR-intensive benchmarks like DocVQA and InfoVQA, a trade-off we analyze next.

\begin{table*}[t]
\centering
\caption{Comparison of OCR, chart, and document understanding performance.
}
\resizebox{.8\linewidth}{!}{
\begin{tabular}{lccccccccc}
\toprule
\textbf{Model} &
\begin{tabular}{@{}c@{}}AI2D\\(w/wo M)\end{tabular} &
\begin{tabular}{@{}c@{}}ChartQA\\(test avg)\end{tabular} &
\begin{tabular}{@{}c@{}}TextVQA\\(val)\end{tabular} &
\begin{tabular}{@{}c@{}}DocVQA\\(test)\end{tabular} &
\begin{tabular}{@{}c@{}}InfoVQA\\(test)\end{tabular} &
\begin{tabular}{@{}c@{}}OCR\\Bench\end{tabular} &
\begin{tabular}{@{}c@{}}SEED-2\\Plus\end{tabular} &
\begin{tabular}{@{}c@{}}CharXiv\\(RQ / DQ)\end{tabular}
\\
\midrule

\multicolumn{9}{l}{\textit{Open-Source Model (Understanding)}} \\
Phi-3.5-Vision-4B~\cite{abdin2024phi3technicalreporthighly} & 77.8 / 87.6 & 81.8 & 72.0 &69.3 & 36.6 & 599 & 62.2 & -\\
Ovis1.6-Gemma2-9B~\cite{lu2024ovis} & 84.4 / - & - & - & - & - & 830 & -& - \\
MiniCPM-V2.6~\cite{yao2024minicpm} & 82.1 / - & 82.4 & 80.1 & 90.8 & 42.6 & 852 & 65.7 & 31.0 / 57.1 \\
Molmo-7B-D~\cite{deitke2025molmo} & - / 93.2 & 84.1 & 81.7 & 92.2 & 72.6 & 694 & - & - \\
InternVL2-8B~\cite{internvl2} & 83.5 / 91.7 & 83.3 & 77.4 & 91.6 & 74.8 & 794 & 67.5 & 31.2 / 56.1 \\
InternVL2.5-8B~\cite{internvl2.5} & 84.5 / 92.8 & 84.8 & 79.1 & 93.0 & 77.6 & 822 & 69.7 & 32.9 / 68.6 \\
Qwen2-VL-7B~\cite{qwen2vl} & 82.0 / 92.1 & 83.0 & 84.3 & 94.5 & 76.4 & 863 & 68.6 & 40.4 / 57.5 \\

\midrule

\multicolumn{9}{l}{\textit{Reasoning Model}} \\
Mulberry-7B~\cite{yao2024mulberry}  & 79.0 / - & 83.9 & - & - & - & - & -& - \\
LLaVA-CoT-11B~\cite{xu2024llavacot} & 85.7 / - & - & - & - & - & - & - & -\\
LLaVA-Reasoner-8B~\cite{llavaReasoner} & 78.5 / - & 83.0 & 71.1 & 81.8 & 51.6 & 637  & - & - \\
Insight-V-8B~\cite{dong2025insight} & - & 81.5 & 76.8 & 91.5 & - & 735 & - & - \\
LlamaV-o1-11B~\cite{thawakar2025llamavo1} & 81.2 / - & - & - & - & - & - & -& -  \\
\midrule
\multicolumn{9}{l}{\textit{Open-Source Model (Unified)}} \\
EVE-7B(HD)~\cite{diao2024unveiling} & - &- & 56.8 & - & - & - & - & -\\
Chameleon-7B~\cite{Chameleon} & 47.4 / - &- & - & - & - & - & 37.3 & -\\
JanusFlow~\cite{JanusFlow}   & 65.3 / 74.3   &64.6 & 55.5 & 28.2 & 13.0 & 569 & 46.4 & 30.0 / 33.8\\
Emu3~\cite{Emu3} & 70.0 / 78.2 & 68.6 & 64.7 & 76.3 & 43.8 & 687 & 44.9 & 24.3 / 20.8  \\
\textbf{OmniBridge (Ours)} & 82.7 / 91.8 & 85.1 & 81.0 & 92.4 & 72.9 & 810 & 68.0 & 40.0 / 56.2 \\

\bottomrule
\end{tabular}
}
\label{tab:vqa-docqa-benchmarks}
\end{table*}

We attribute this performance trade-off to a strategic shift in the model's focus from low-level perception to high-level abstraction, driven by two factors. First, our training strategy, which emphasizes multi-step logical reasoning for mathematics, likely biases the model's representational capacity towards abstract semantics. This inadvertently de-prioritizes the fine-grained visual-text alignment and positional awareness critical for OCR and document layout parsing. Second, this reflects a classic challenge in unified modeling: a trade-off between abstract reasoning and perceptual fidelity. We posit that overcoming this will require future work in areas like adaptive, task-specific modules or dynamic capacity allocation to achieve more balanced cross-task generalization.

\begin{table}[htbp]
\centering
\caption{Comparison of multi-image understanding performance. 
}
\resizebox{.8\linewidth}{!}{%
\begin{tabular}{lcccc}
\toprule

\textbf{Model} &
\begin{tabular}{@{}c@{}}BLINK\\(val)\end{tabular} &
\begin{tabular}{@{}c@{}}Mantis\\Eval\end{tabular} &
\begin{tabular}{@{}c@{}}Muir\\Bench\end{tabular} &
\begin{tabular}{@{}c@{}}MMT\\(val)\end{tabular} \\

\midrule
\multicolumn{5}{l}{\textit{Open-Source Model (Understanding)}} \\
Phi-3.5-Vision-4B~\cite{abdin2024phi3technicalreporthighly} & 58.3 & -    & -    & 53.6 \\
MiniCPM-V2.6~\cite{yao2024minicpm}      & 53.0 & 69.0 & -    & 60.8 \\
InternVL2-8B~\cite{internvl2}      & 50.9 & 65.4 & 48.7 & 60.0 \\
InternVL2.5-8B~\cite{internvl2.5}    & 54.8 & 67.5 & 51.1 & 62.3 \\
Qwen2-VL-7B~\cite{qwen2vl}       & 53.2 & 66.8 & -    & 64.0 \\
\midrule
\multicolumn{5}{l}{\textit{Open-Source Model (Unified)}} \\
Show-o~\cite{Show-o}            & 25.1 & -    & -    & -    \\
EVE-7B (HD)~\cite{diao2024unveiling}       & -    & -    & -    & 62.6 \\
JanusFlow~\cite{JanusFlow}         & 41.7 & 30.9 & 25.7 & 49.0 \\
Emu3~\cite{Emu3}           & 31.6 & 31.2 & 27.0    & 60.3 \\
\textbf{OmniBridge (Ours)} & 53.3 & 71.4 & 43.8 & 62.5 \\
\bottomrule
\end{tabular}%
}
\label{tab:exp:multi-image}
\end{table}

\subsubsection{Multi-Image Understanding}



OmniBridge demonstrates exceptional multi-image reasoning capabilities, achieving state-of-the-art performance among unified open-source models as shown in Table~\ref{tab:exp:multi-image}. Specifically, it achieves 71.4 on Mantis-Eval and 62.5 on MMT-Bench, substantially outperforming prior unified models like Emu3 and JanusFlow. More notably, OmniBridge's performance matches or surpasses that of several powerful, specialized understanding-only models, including its own backbone Qwen2-VL-7B and InternVL2.5-8B. This result is significant, challenging the common assumption that unifying generation and understanding necessitates a compromise in comprehension performance. We attribute this success to our training strategy: by enhancing a strong comprehension-centric base with our R1 distillation and RL phases, OmniBridge achieves a synergistic effect where the model's reasoning capabilities are elevated, rather than diluted, by the integration of multiple tasks.

\subsubsection{Real-World Comprehension}



OmniBridge demonstrates robust performance in complex real-world settings, achieving consistently strong results across the RealWorldQA, MME-RW (EN), and R-Bench benchmarks (Table~\ref{tab:exp:realworld}). It not only outperforms other unified open-source models but also approaches or surpasses several dedicated understanding models, attaining high scores of 66.1 on R-Bench and 64.5 on RealWorldQA. We attribute this resilience to real-world ambiguity and variation directly to our R1-style distillation strategy. This strategy injects step-wise, human-like reasoning supervision into the model, compelling it to adopt a "reason-then-answer" approach. By learning to decompose and explain complex visual-textual cues, OmniBridge becomes inherently more robust against the nuances encountered in real-world inputs.

\begin{table}[ht]
\centering
\caption{Comparison of real-world performance.
}
\resizebox{.9\linewidth}{!}{%
\begin{tabular}{lccc}
\toprule

\textbf{Model} & \textbf{RealWorld (QA)} & \textbf{MME-RW (EN)} & \textbf{R-Bench dis} \\
\midrule
\multicolumn{4}{l}{\textit{Open-Source Model}} \\
Phi-3.5-Vision-4B~\cite{abdin2024phi3technicalreporthighly} & 53.6 & -    & 55.5 \\
MiniCPM-V2.6~\cite{yao2024minicpm}      & 65.0 & -    & -    \\
InternVL2-8B~\cite{internvl2}      & 64.4 & 53.5 & 67.9 \\
InternVL2.5-8B~\cite{internvl2.5}    & 70.1 & 59.1 & 70.1 \\
Qwen2-VL-7B~\cite{qwen2vl}       & 58.2 & 56.5 & 64.0 \\
\midrule
\multicolumn{4}{l}{\textit{Open-Source Model (Unified)}} \\
Chameleon-7B~\cite{Chameleon}      & 39.6 & -    & -    \\
Emu3~\cite{Emu3}           & 57.3 & -    & 54.5 \\
JanusFlow~\cite{JanusFlow}         & 55.8 & 32.0 & 58.8 \\
\textbf{OmniBridge (Ours)} & 64.5 & 57.3   & 66.1 \\
\bottomrule
\end{tabular}%
}
\label{tab:exp:realworld}
\end{table}

\subsubsection{Comprehensive Multimodal Evaluation}


On comprehensive benchmarks, OmniBridge establishes a new state-of-the-art for open-source unified models, as detailed in the left part of Table~\ref{tab:exp:cm-he-benchmarks}. This is best exemplified by its leading MME score of 2352.9, which significantly surpasses prior unified baselines like Emu3 and JanusFlow. Furthermore, it achieves highly competitive scores on MMStar (58.9) and MMBench (79.1), substantially narrowing the performance gap to powerful, dedicated understanding models like InternVL2.5-8B. Collectively, these results validate the efficacy of the introduced R1 distillation strategy. By enhancing the core perception and reasoning of the base model, our approach successfully bridges the gap, enabling a unified model to achieve performance on par with its task-specific counterparts.

\begin{table}[t]
\centering
\caption{
Comparison of comprehensive multimodal understanding and hallucination performance.
}
\resizebox{\linewidth}{!}{%
\begin{tabular}{lccc|cc}
\toprule
\textbf{Model} &
\begin{tabular}{@{}c@{}}MME\\(sum)\end{tabular} &
MMStar &
\begin{tabular}{@{}c@{}}MMBench\\(EN)\end{tabular} &
\begin{tabular}{@{}c@{}}HallBench\\(avg)\end{tabular} &
\begin{tabular}{@{}c@{}}POPE\\(avg)\end{tabular}  \\
\midrule
\multicolumn{6}{l}{\textit{Open-Source Model (Understanding)}} \\

Phi-3.5-Vision-4B~\cite{abdin2024phi3technicalreporthighly}   & -      & 47.5 & 76.0  & 40.5    & -    \\
MiniCPM-V2.6~\cite{yao2024minicpm}        & 2348.4 & 60.0 & 81.5  & 48.1 & 87.3 \\
InternVL2-8B~\cite{internvl2}        & 2210.3 & 54.2 & 81.7  & 45.2 & 86.9 \\
InternVL2.5-8B~\cite{internvl2.5}      & 2344.1 & 62.8 & 84.6  & 50.1 & 90.6 \\
Qwen2-VL-7B~\cite{qwen2vl}         & 2324.6 & 60.7 & 83.0  & 50.6 & 87.1 \\

\midrule
\multicolumn{6}{l}{\textit{Reasoning Model}} \\
LLaVA-CoT-11B~\cite{xu2024llavacot} & - & 57.6 &   & 47.8 & -  \\
LLaVA-Reasoner-8B~\cite{llavaReasoner} & - & 54.0 & - & - & -  \\
Insight-V-8B~\cite{dong2025insight} & 2069 & 58.6 & - & - & -   \\
LlamaV-o1-11B~\cite{thawakar2025llamavo1} & - & 59.5 &   & 63.5 & -  \\
Mulberry-7B~\cite{yao2024mulberry} & 2396 & 61.3 &   & 54.1 & -  \\

\midrule
\multicolumn{6}{l}{\textit{Open-Source Model (Unified)}} \\
Chameleon-7B~\cite{Chameleon}       &  123.9      & 19.3 &   & 42.5 & 17.8   \\
Show-o~\cite{Show-o}             & 1182.7 & - & - & - & 73.8   \\

SEED-X~\cite{SEED-X}             & 1435.7 & - & - & - & -   \\
EVE-7B (HD)~\cite{diao2024unveiling}        & 1305.7      & - & - & - & 85.0   \\
VILA-U~\cite{VILA-U}             & 1401.8 & - & - & - & 85.8   \\
Janus~\cite{Janus}              & 1338.0 & - & - & - & 87.0  \\
JanusFlow~\cite{JanusFlow}          & 1333.1 & 46.7 & 66.3  & 34.0 & 85.3 \\
Emu3~\cite{Emu3}               &   1590.7    & 46.6 & 63.4 & 27.2 & 82.9   \\
\textbf{OmniBridge (Ours)}  & 2352.0 & 58.9 & 79.1  & 53.4 & 90.9   \\

\bottomrule
\end{tabular}
}
\label{tab:exp:cm-he-benchmarks}
\end{table}

\begin{figure*}[hbtp]
\centering
\includegraphics[width=\linewidth]{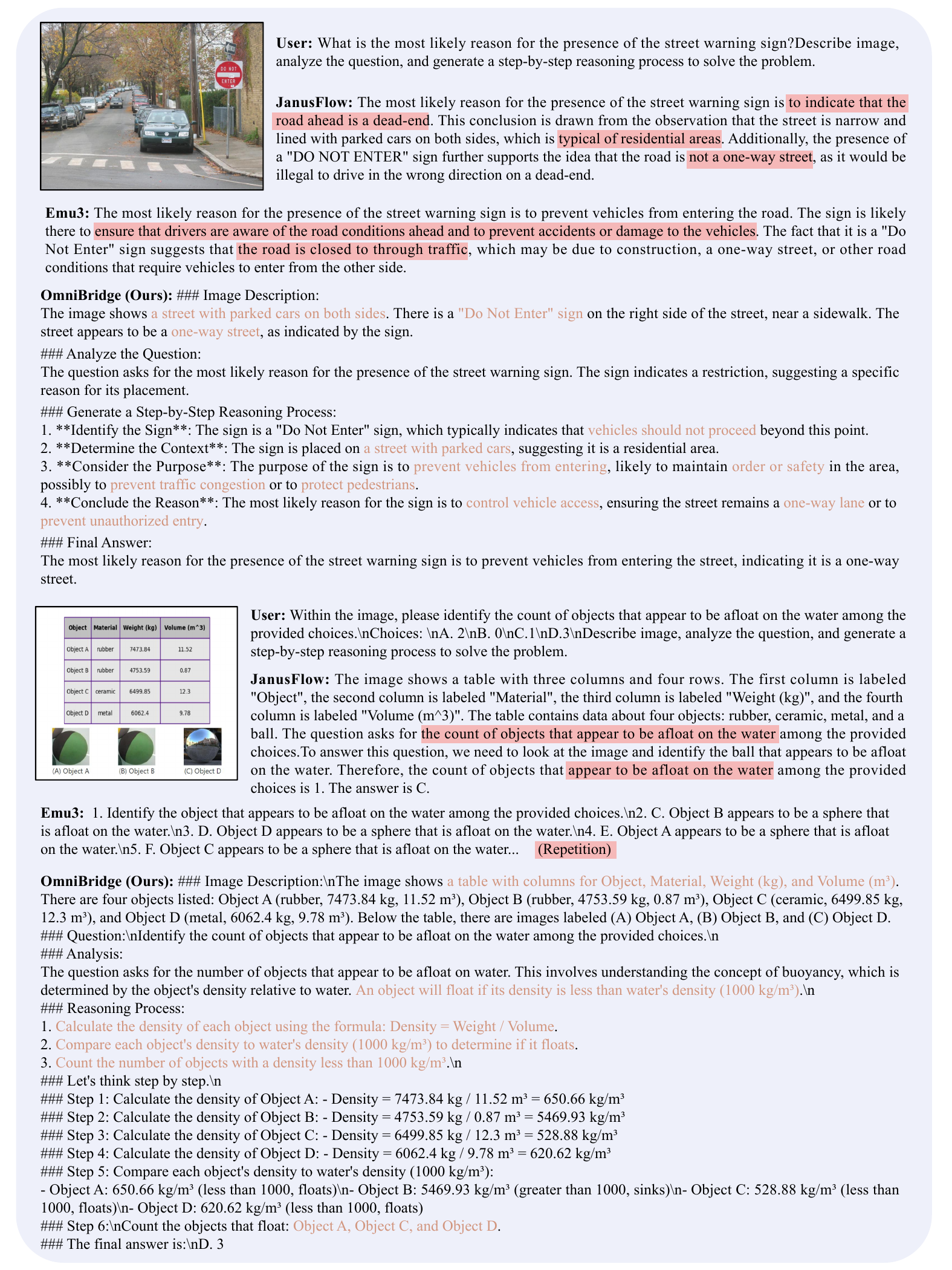}
\caption{\textbf{Qualitative results of multimodal understanding on multi-hop reasoning}. We compare the response with Emu3~\cite{Emu3} and JanusFlow~\cite{JanusFlow}. We emphasize the key-points in the
response. Best viewed on screen.}
\label{fig:understanding case}
\end{figure*}

\subsubsection{Multimodal Hallucination Evaluation}
OmniBridge demonstrates exceptional robustness against multimodal hallucinations, achieving strong performance on both HallucinationBench and POPE, as detailed in the right part of Table~\ref{tab:exp:cm-he-benchmarks}. With scores of 53.4 on HallucinationBench and 90.9 on POPE, it not only outperforms most unified competitors like Emu3 and JanusFlow but also exhibits comparable or superior hallucination resistance to specialized understanding models such as InternVL2.5-8B. This latter comparison is particularly telling, as it suggests that the model's low hallucination rate is a direct byproduct of its enhanced reasoning capabilities. We posit that the introduced R1 distillation strategy, by instilling a step-by-step thinking process, inherently promotes factual grounding, compelling the model to verify its outputs through logical decomposition rather than relying on superficial pattern association.

\begin{table*}[htbp]
\centering
\caption{\textbf{Performances on GenEval benchmark.} ``Gen.'' denotes ``generation'' and ``Unified'' denotes unified understanding and generation models. Models using learnable queries instead of conditional embeddings are signed with †.}
\resizebox{0.8\linewidth}{!}{
\begin{tabular}{llccccccc}
\toprule
\textbf{Type} & \textbf{Method} & \textbf{Single Obj.} & \textbf{Two Obj.} & \textbf{Counting} & \textbf{Colors} & \textbf{Position} & \textbf{Color Attr.} & \textbf{Overall$\uparrow$} \\ 
\midrule
\multirow{10}{*}{\textbf{\textit{Gen. Only}}} & LlamaGen \cite{LLamaGen}  & 0.71 & 0.34 & 0.21 & 0.58 & 0.07 & 0.04 & 0.32 \\
 & LDM \cite{LDM} & 0.92 & 0.29 & 0.23 & 0.70 & 0.02 & 0.05 & 0.37 \\
 & SDv1.5 \cite{LDM} & 0.97 & 0.38 & 0.35 & 0.76 & 0.04 & 0.06 & 0.43 \\
 & PixArt-$\alpha$ \cite{PixArt-alpha} & 0.98 & 0.50 & 0.44 & 0.80 & 0.08 & 0.07 & 0.48 \\
 & SDv2.1 \cite{LDM} & 0.98 & 0.51 & 0.44 & 0.85 & 0.07 & 0.17 & 0.50 \\
 & DALL-E 2 \cite{DALL-E2} & 0.94 & 0.66 & 0.49 & 0.77 & 0.10 & 0.19 & 0.52 \\
 & SDXL \cite{SDXL} & 0.98 & 0.74 & 0.39 & 0.85 & 0.15 & 0.23 & 0.55 \\
 & IF-XL \cite{IF-XL} & 0.97 & 0.74 & 0.66 & 0.81 & 0.13 & 0.35 & 0.61 \\
 & DALL-E 3 \cite{DALL-E3}  & 0.96 & 0.87 & 0.47 & 0.83 & 0.43 & 0.45 & 0.67 \\
 & HunyuanDiT \cite{HunyuanDiT}  & 0.96 & 0.74 & 0.59 & 0.82 & 0.10 & 0.42 & 0.59 \\
 
\midrule
\multirow{10}{*}{\textit{\textbf{Unified}}} & Chameleon \cite{Chameleon}  & - & - & - & - & - & - & 0.39 \\
 & LWM \cite{LWM} & 0.93 & 0.41 & 0.46 & 0.79 & 0.09 & 0.15 & 0.47 \\
 & SEED-X \cite{SEED-X} & 0.97 & 0.58 & 0.26 & 0.80 & 0.19 & 0.14 & 0.49 \\
 & Show-o \cite{Show-o} & 0.95 & 0.52 & 0.49 & 0.82 & 0.11 & 0.28 & 0.53 \\
 & Emu3 \cite{Emu3} & 0.98 & 0.71 & 0.34 & 0.81 & 0.15 & 0.21 & 0.54 \\
 & Janus \cite{Janus} & 0.97 & 0.68 & 0.30 & 0.84 & 0.46 & 0.42 & 0.61 \\
 & Transfusion \cite{Transfusion}  & - & - & - & - & - & - & 0.63 \\
 & JanusFlow \cite{JanusFlow} & 0.97 & 0.59 & 0.45 & 0.83 & 0.53 & 0.42 & 0.63 \\
 & \textbf{OmniBridge† (Ours)} & 0.96  & 0.65 & 0.49 & 0.80 & 0.19 & 0.28 & 0.56 \\
 & \textbf{OmniBridge (Ours)} & 0.96  & 0.79 & 0.65 & 0.74 & 0.20 & 0.31 & 0.61 \\
\bottomrule
\end{tabular}
}
\label{tab:exp:GenEval_comparison}
\end{table*}

\subsubsection{Qualitative Results} 

Figure~\ref{fig:understanding case} qualitatively demonstrates OmniBridge's superior complex reasoning capabilities in two challenging scenarios. In the first commonsense reasoning case, competing models misinterpret a "Do Not Enter" sign: JanusFlow~\cite{JanusFlow} provides a self-contradictory explanation, while Emu3~\cite{Emu3}'s response is overly broad. In stark contrast, OmniBridge exhibits structured, human-like reasoning. It leverages the visual context of a "narrow residential street" to correctly infer the sign's purpose—enforcing a "one-way lane"—showcasing its strong situational awareness.

This advantage is even more pronounced in the second task, which requires quantitative scientific reasoning. Here, competing models fail entirely, offering speculative answers devoid of any calculation.  OmniBridge, however, adopts a rigorous scientific methodology: it identifies the core physical concept ("density"), formulates the necessary equation, extracts data from the table, and executes a step-by-step calculation.  This transparent, multi-hop process, which seamlessly integrates visual parsing, knowledge retrieval, and mathematical operations, proves its robust capability for solving complex, structured problems.

\subsection{Image Generation}

\subsubsection{Quantitative Evaluation}

Our quantitative evaluations on GenEval and DPG-Bench (Tables~\ref{tab:exp:GenEval_comparison} and \ref{tab:exp:DPG-Bench}) confirm that OmniBridge achieves performance competitive with state-of-the-art models, despite its high data efficiency. On GenEval, our model scores 0.61, placing it on par with leading unified models like JanusFlow (0.63) and significantly outperforming Emu3 (0.54). A detailed breakdown reveals particular strengths in compositional tasks such as "Two Objects" (0.79) and "Counting" (0.65). Crucially, our variant OmniBridge†, which operates on learnable queries processed by the BiTransformer, achieves a strong score of 0.56 without any explicit text conditioning, demonstrating effective latent space control. On the more instruction-rich DPG-Bench, OmniBridge scores 78.93\%, surpassing strong baselines like Playground v2.5~\cite{Playgroundv2.5} and remaining competitive with Emu3.

\begin{table}[h]
\centering
\caption{Performances on DPG-bench benchmark.}
\resizebox{\linewidth}{!}{
\begin{tabular}{lcccccc}
\toprule
\textbf{Method} & \textbf{Global} & \textbf{Entity} & \textbf{Attribute} & \textbf{Relation} & \textbf{Other} & \textbf{Overall$\uparrow$} \\ 
\midrule
\multicolumn{7}{l}{\textit{\textbf{Gen. Only}}} \\
 SDv1.5 \cite{LDM}                      & 74.63 & 74.23 & 75.39 & 73.49 & 67.81 & 63.18 \\ 
 PixArt-$\alpha$ \cite{PixArt-alpha}   & 74.97 & 79.32 & 78.60 & 82.57 & 76.96 & 71.11 \\ 
 Lumina-Next \cite{Lumina-Next}         & 82.82 & 88.65 & 86.44 & 80.53 & 81.82 & 74.63 \\ 
 SDXL \cite{SDXL}                       & 83.27 & 82.43 & 80.91 & 86.76 & 80.41 & 74.65 \\ 
 Playground v2.5 \cite{Playgroundv2.5}  & 83.06 & 86.59 & 81.20 & 84.08 & 83.50 & 75.47 \\ 
 Hunyuan-DiT \cite{HunyuanDiT}          & 84.59 & 80.59 & 88.01 & 74.36 & 86.41 & 78.87 \\ 
 PixArt-$\Sigma$ \cite{Pixart-sigma}    & 86.89 & 89.94 & 88.94 & 86.59 & 87.68 & 80.54 \\ 
\midrule
\multicolumn{7}{l}{\textit{\textbf{Unified}}} \\
 Show-o~\cite{Show-o}  & 79.33 & 75.44 & 78.02 & 84.45 & 60.80 & 67.27 \\

 Emu3 \cite{Emu3}                       & 85.21 & 86.68 & 86.84 & 90.22 & 83.15 & 80.60 \\ 
 JanusFlow \cite{JanusFlow}             & 87.03 & 87.31 & 87.39 & 89.79 & 88.10 & 80.09 \\

 TokenFlow-XL~\cite{qu2025tokenflow}  & 78.72 & 79.22 & 81.29  & 85.22 & 71.20 & 73.38 \\ 
  \textbf{OmniBridge† (Ours)} & 76.90 & 80.44 & 80.41 & 86.03 & 57.60 & 71.37 \\
 \textbf{OmniBridge (Ours)}              &   80.55  &  85.85  &  84.94  & 90.44  & 74.0  & 78.93  \\ 
\bottomrule
\end{tabular}
}
\label{tab:exp:DPG-Bench}
\end{table}

This strong performance is particularly noteworthy given two key factors. First, data efficiency: OmniBridge's generation module was trained on only 7K samples, a fraction of the data used by competing models. Second, architectural integrity: our approach does not modify the base diffusion model's decoder. This indicates that the performance gains stem directly from our novel alignment strategy rather than architectural enhancements to the generator itself. Our training process, which enriches short captions into detailed descriptions, likely explains the strong performance on DPG-Bench's dense prompts. While GenEval's shorter prompts may not fully exploit this strength, our model's competitive score there nonetheless validates its robust and versatile generative foundation.

\begin{figure*}[hbtp]
\centering
\includegraphics[width=\linewidth]{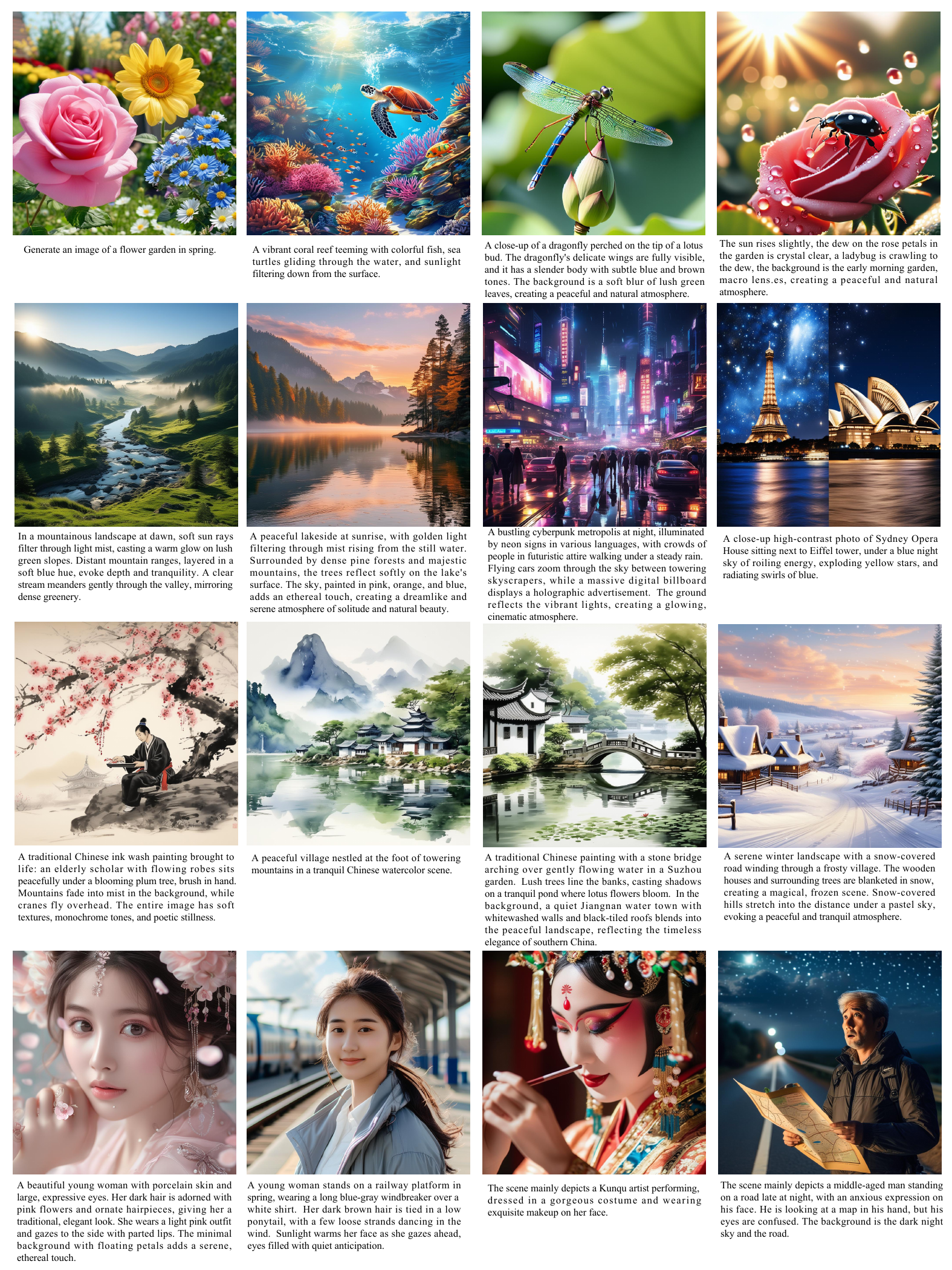}
\caption{\textbf{Qualitative results of text-to-image generation of OmniBridge}. OmniBridge can generate high-quality images that are semantically consistent with text prompts. We upsample the images to 1024 × 1024 for better visualization.}
\label{fig:generation case}
\end{figure*}

\subsubsection{Qualitative Results}

\begin{figure*}[hbtp]
\centering
\includegraphics[width=.9\linewidth]{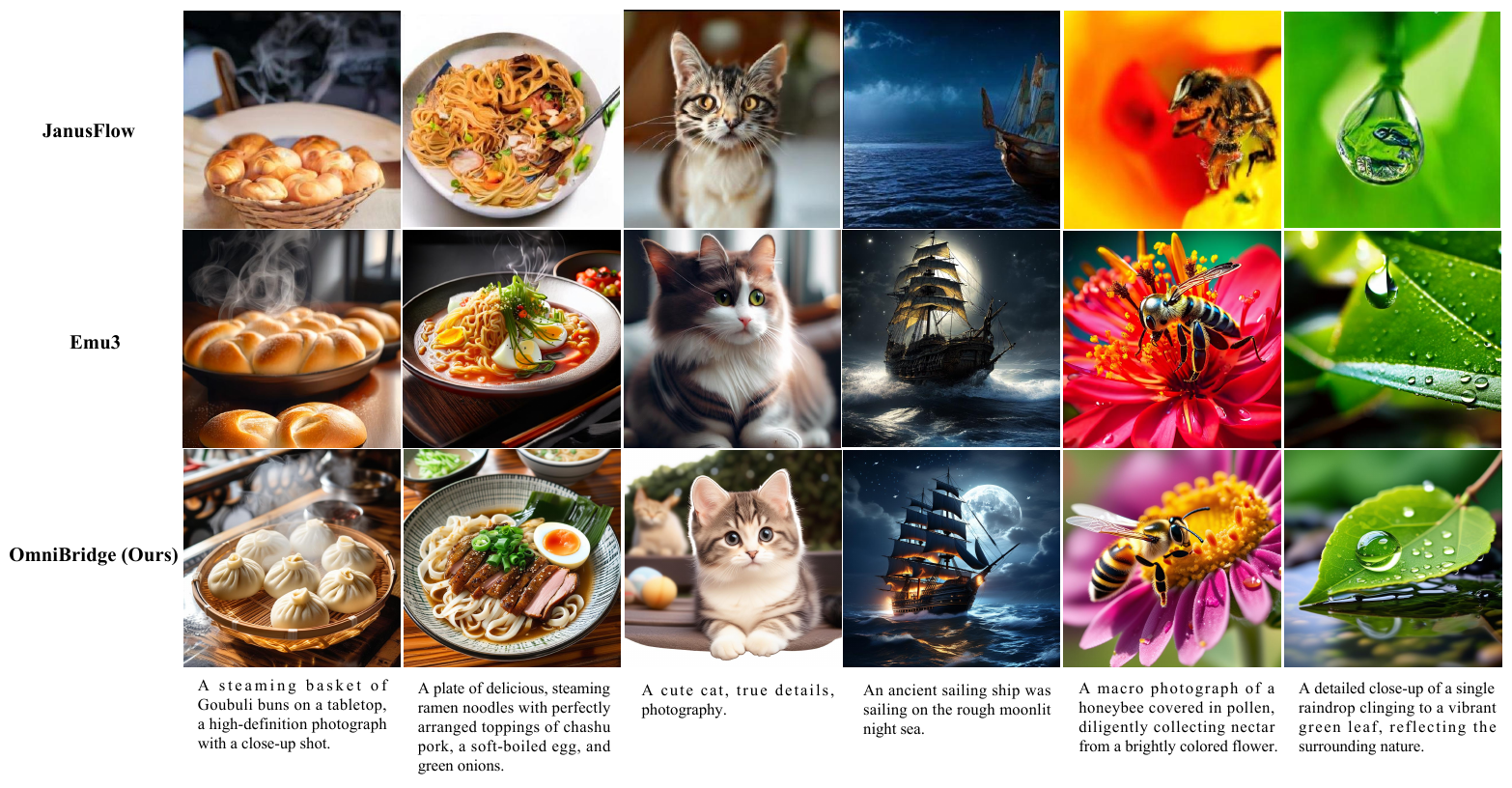}
\caption{\textbf{Qualitative comparisons of visual generation with JanusFlow and Emu3}. The images generated by OmniBridge show better consistency with the user’s prompts. }
\label{fig:Image cases}
\end{figure*}

We qualitatively evaluate OmniBridge's generation capabilities in Figures~\ref{fig:generation case} and \ref{fig:Image cases}. A direct comparison against SOTA models in Figure~\ref{fig:Image cases} highlights OmniBridge's superior prompt fidelity and semantic accuracy. While JanusFlow~\cite{JanusFlow} struggles with both image quality and basic semantic correctness (e.g., producing spaghetti instead of ramen), Emu3~\cite{Emu3} presents a more subtle failure mode. Despite generating high-fidelity images, Emu3 often deviates from critical details in the prompt, such as placing buns on a plate instead of the requested ``basket'' or creating visual artifacts like a ``third ear'' on a cat. In stark contrast, OmniBridge correctly renders all specified details—from the basket for the buns to the cat's correct anatomy—demonstrating superior semantic consistency. Crucially, this higher degree of prompt fidelity is achieved with significant efficiency gains, as OmniBridge generates images several times faster than Emu3.



\subsection{Image Editing}

\subsubsection{Quantitative Evaluation} 



OmniBridge's remarkable data efficiency and strong performance in image editing are validated on the ImgEdit-Bench~\cite{ye2025imgedit}, with results in Table~\ref{tab:exp:imgedit}. Despite its BiTransformer being trained on only 5,500 samples from MagicBrush~\cite{zhang2023magicbrush}—a tiny fraction of the data used by specialized or commercial models—OmniBridge demonstrates highly competitive results. It significantly outperforms its training-set provider, MagicBrush (Overall 2.28 vs. 1.90), and surpasses other specialized editors like Instruct-P2P~\cite{InstructBLIP} and AnyEdit~\cite{yu2025anyedit} on several tasks.

Most notably, OmniBridge achieves a SOTA score of 4.34 in the "Remove" task, surpassing even the significantly larger GPT-4o (3.66). Furthermore, it outperforms the non-commercial baselines in the "Extract" (2.30) and "Action" (3.64) tasks. These results powerfully demonstrate that our latent space alignment strategy can achieve exceptional generalization and data efficiency, enabling strong performance with minimal task-specific training.

\begin{table*}[htbp]
\centering
\caption{\textbf{Comparison results on ImgEdit-Bench~\cite{zhang2023magicbrush}}. ``Overall" is calculated by averaging all scores across tasks.}
\resizebox{0.95\linewidth}{!}{%
\begin{tabular}{lcccccccccccc}
\toprule
\textbf{Model} & \# Train Samples  & Add & Adjust & Extract & Replace & Remove & Background & Style & Hybrid & Action & \textbf{Overall} $\uparrow$ \\
\midrule
GPT-4o~\cite{hurst2024gpt} & - & \textbf{4.61} & \textbf{4.33} & \textbf{2.90} & \textbf{4.35} & 3.66 & \textbf{4.57} & \textbf{4.93} & \textbf{3.96} & \textbf{4.89} & \textbf{4.20} \\
\midrule
MagicBrush~\cite{zhang2023magicbrush} & 470K & 2.84 & 1.58 & 1.51 & 1.97 & 1.58 & 1.75 & 2.38 & 1.62 & 1.22 & 1.90 \\
Instruct-P2P~\cite{brooks2023instructpix2pix} & 450K & 2.45 & 1.83 & 1.44 & 2.01 & 1.50 & 1.44 & 3.55 & 1.22 & 1.46 & 1.88 \\
AnyEdit~\cite{yu2025anyedit} & 2.5M & 3.18 & 2.95 & 1.88 & 2.47 & 2.23 & 2.24 & 3.68 & 2.12 & 2.65 & 2.45 \\
UltraEdit~\cite{UltraEdit} & 3M & 3.60 & 3.06 & 2.24 & 2.58 & 2.45 & \textbf{3.76} & 1.91 & \textbf{2.98} & 3.22 & 2.80 \\
OmniGen~\cite{OmniGen} & 4.13M & 3.47 & 3.04 & 2.06 & 2.94 & 2.43 & 2.91 & 3.76 & 2.46 & 2.70 & 2.96 \\
Step1X-Edit~\cite{Step1X-Edit} & 20M & \textbf{3.88} & 3.14 & 2.30 & 3.06 & 2.41 & 3.16 & 3.63 & 2.42 & 3.23 & 3.06 \\
ICEdit~\cite{ICEdit} & 50K & 3.58 & 3.39 & 1.76 & 3.30 & 2.51 & 3.00 & 3.44 & 2.32 & 3.08 & 2.96 \\
BAGEL~\cite{BAGEL} & 5T tokens & 3.25 & 3.13 & 1.64 & 3.18 & 2.44 & 3.32 & \textbf{4.49} & 2.38 & 3.17 & 3.14 \\
\textbf{OmniBridge (Ours)} & 5.5K & 1.05 & 3.09 & \textbf{2.30} & 1.76 & \textbf{4.34} & 1.89 & 1.20 & 1.28 &\textbf{3.64} & 2.28 \\
\bottomrule
\end{tabular}
}
\label{tab:exp:imgedit}
\end{table*}

\subsubsection{Qualtative Results}

\begin{figure}[htbp]
\centering
\includegraphics[width=.95\linewidth]{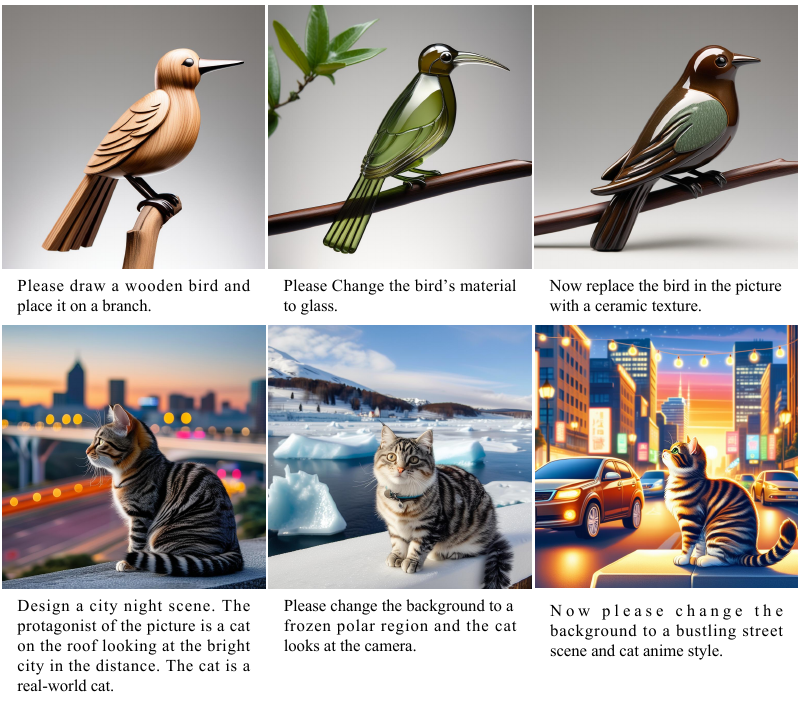}
\caption{Qualtative Results of image editing.}
\label{fig:Image_editing}
\end{figure}

Figure~\ref{fig:Image_editing} provides qualitative evidence of OmniBridge's proficiency in complex, instruction-based image editing, where it performs major semantic modifications while preserving overall coherence. For instance, the top row shows the model successfully changing a bird's material sequentially from wood to glass, and then to ceramic, all while maintaining the object's original pose and the scene's composition. The bottom row showcases a more advanced, multi-turn editing capability. The model first executes a complex instruction—replacing a background and changing the subject's gaze simultaneously. Building on this, it then flawlessly applies a subsequent edit, transforming the background again while rendering the cat in a distinct "anime style." This ability to handle sequential, compositional edits involving attributes, backgrounds, and artistic style highlights its robust instruction-following capacity.

Nevertheless, we note that the model's proficiency in highly detailed, pixel-level editing is an area for development. This is an inherent limitation of our approach, as the OmniBridge is optimized for high-level semantic representation, making fine-grained spatial manipulation challenging. Future work could focus on integrating more granular control mechanisms to bridge this gap between semantic and pixel-level editing.

\subsection{Multimodal Retrieval}



OmniBridge demonstrates state-of-the-art multimodal retrieval performance on the Flickr30K benchmark~\cite{plummer2015flickr30k}, as detailed in Table~\ref{tab:retrieval-performance}. Achieving an overall average score of 92.5, it surpasses all listed baselines, including highly competitive specialized models like InternVL-C (92.2) and OpenCLIP-G (91.4). A detailed breakdown reveals particular strength in text-to-image retrieval (T2I), where it scores 84.8 at R@1, outperforming most competitors. Furthermore, its recall at higher ranks is exceptional, with R@5 and R@10 scores consistently approaching the theoretical maximum across all tasks. This superior performance validates our architectural design. We attribute this success to our BiTransformer module, which performs fine-grained alignment on the rich hidden states of the base LLM. By effectively mapping image-text pairs into a coherent shared latent space, this mechanism enables a more discriminative and robust cross-modal representation.

\begin{table*}[t]
\centering
\caption{Comparison of zero-shot image-text retrieval performance. We evaluate the retrieval capability using the Flickr30K~\cite{plummer2015flickr30k}. R@k denotes Recall@k (\%).}
\resizebox{.85\linewidth}{!}{%
\begin{tabular}{lcccccccccc}
\toprule
\multirow{2}{*}{\textbf{Method}} &
\multicolumn{9}{c}{\textbf{Flickr30K (test set)}} &
\multirow{2}{*}{\textbf{Avg.}} \\
\cmidrule(lr){2-10} 
& \multicolumn{3}{c}{Image $\rightarrow$ Text} & \multicolumn{3}{c}{Text $\rightarrow$ Image} & \multicolumn{3}{c}{Text $\rightarrow$ Text}
 \\
& R@1 & R@5 & R@10 & R@1 & R@5 & R@10 & R@1 & R@5 & R@10  \\
\midrule
OpenCLIP-H~\cite{OpenCLIP}          & 88.2 & 98.3 & 99.3 & 76.0 & 93.0 & 96.1 & 71.4 & 87.3 & 91.8 & 89.0 \\
OpenCLIP-g~\cite{OpenCLIP}         & 86.6 & 98.2 & 98.9 & 73.6 & 92.3 & 95.6 & 70.6 & 86.5 & 91.2 & 88.2 \\
OpenCLIP-G~\cite{OpenCLIP}          & 92.7 & 99.4 & 99.7 & 78.9 & 94.6 & 97.0 & 72.8 & 88.5 & 92.6 & 91.4 \\
EVA-01-CLIP-g+~\cite{sun2023eva}          & 91.6 & 99.3 & 99.7 & 78.9 & 94.5 & 96.9  & 68.7 & 84.4 & 88.6 & 89.2 \\

EVA-02-CLIP-E+~\cite{sun2023eva}          & 93.8 & 99.4 & 99.8 & 78.8 & 94.2 & 96.8 & 74.5 & 90.1 & 93.8 &  91.2\\
BLIP-2~\cite{li2023blip}       & 89.3 & 99.7 & 99.9 & 81.5 & 96.0 & 98.0 & 63.4 & 81.7 & 87.3 & 88.5  \\
BLIP-2-G~\cite{li2023blip}       & 90.6 & 99.7 & 99.9 & 85.7 & 97.6 & 99.0 & 74.2 & 89.5 & 93.4 & 92.2 \\
InternVL-C~\cite{chen2024internvl}               & 94.0  & 99.3 & 99.7 & 81.3 & 95.7 & 97.8 &  74.7 & 90.0 & 94.1 &  91.8 \\
BGE-M3~\cite{multi2024m3}               &  - & - & - & - & - & - &  66.8 & 85.9 & 91.2 & -  \\
BGE-VL-base~\cite{zhou2024megapairs}               & 75.3 & 92.8 & 96.3 & 71.0 & 90.5 & 94.3 &  61.7 & 80.9 & 87.1 &  83.3 \\
BGE-VL-MLLM-S1~\cite{zhou2024megapairs}               & 69.8  & 89.2 & 93.5 & 71.2 & 90.4 & 94.5 &  51.5 & 68.9 & 74.9 &  78.2 \\

\textbf{OmniBridge (Ours)}     & 91.8 & 99.5 & 99.9  &  84.8  & 98.7 & 99.6  & 73.2 & 90.3 & 94.7 & 92.5

\\

\bottomrule
\end{tabular}%
}
\label{tab:retrieval-performance}
\end{table*}

\subsection{Ablation Studies}

\subsubsection{Impact of Latent Space Alignment}

We present an ablation study in Table~\ref{tab:exp:with-rl} to validate the advantage of our latent space alignment, demonstrating how latent space alignment enhances its target capabilities without mutual interference. 

First, we analyze the impact of Stage 1 on multimodal understanding. OmniBridge shows dramatic improvements over the Qwen2VL-7B backbone on understanding benchmarks, such as M3CoT (56.3 → 62.2), Mantis Eval, and RealWorldQA. This confirms that our combination of CoT distillation and StepGRPO successfully enhances the LLM's core reasoning abilities. 
Crucially, the effectiveness of our decoupled design is evidenced by the generation and retrieval scores. The results for GenEval, DPG Bench, and Flickr30K are exclusively products of Stage 2 (Latent Space Alignment), which is conducted while the LLM backbone is frozen. 
The strong performance on these tasks validates the efficacy of our latent space alignment. 
Since StepGRPO is a Stage 1 modification to the LLM, the fact that it does not alter the Stage 2 results confirms that the two stages operate independently, successfully mitigating multi-task interference.

Collectively, this demonstrates that the latent space alignment enables OmniBridge to achieve high performance across understanding, generation, and retrieval within a single framework, validating the core hypothesis of our work.

\begin{table}[h]
\centering
\caption{
Ablation study of Latent Space Alignment over OmniBridge.
}
\resizebox{\linewidth}{!}{%
\begin{tabular}{lcccc|cc|c}
\toprule
\textbf{Model} &
M3CoT   &
\begin{tabular}{@{}c@{}}Mantis\\Eval\end{tabular} &
RealWorldQA &
\begin{tabular}{@{}c@{}}HallBench\\(avg)\end{tabular} &
GenEval &
\begin{tabular}{@{}c@{}}DPG\\Bench\end{tabular} &
\begin{tabular}{@{}c@{}}Flickr30K\\Avg.\end{tabular} 
\\
\midrule
Qwen2VL-7B~\cite{qwen2vl}       &   56.3  & 66.8  & 58.2 & 50.6  & - &  - & -\\
\midrule
JanusFlow~\cite{JanusFlow}       &  43.8  & 30.9 & 55.8 & 34.0  & 63.1 & 80.1 & -\\
Emu3~\cite{Emu3}          &   43.1   &   31.2    & 57.3 & 27.0 & 54.3 & 80.6  & - \\
\midrule
\begin{tabular}{@{}l@{}}OmniBridge\\-w/o StepGRPO \end{tabular}& 60.2 &  72.8 & 62.9 & 64.8  & 61.4 & 78.9  & 92.5 \\
OmniBridge  & 62.2 & 71.4 & 64.5 & 53.4  & 61.4 & 78.9 & 92.5  \\

\bottomrule
\end{tabular}
}
\label{tab:exp:with-rl}
\end{table}

\begin{figure}[h]
\centering
\includegraphics[width=.95\linewidth]{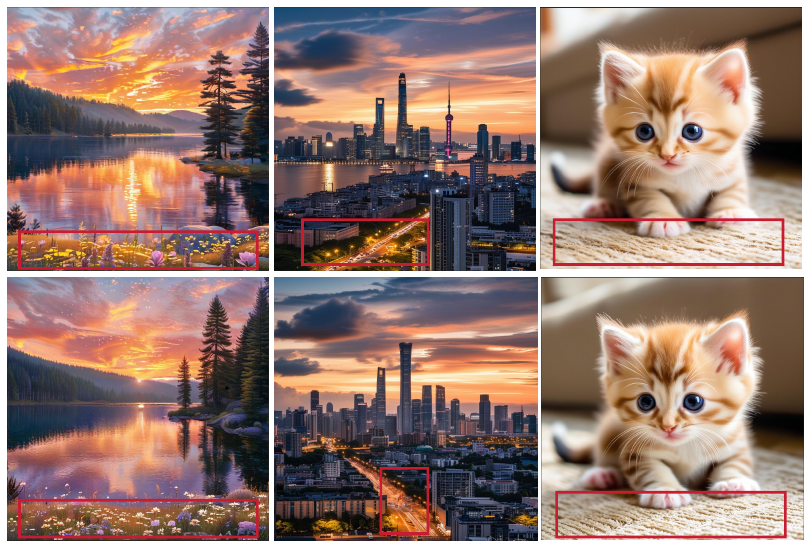}
\caption{Ablation study on the contribution of the Bidirectional Transformer. The results compare a unidirectional variant with our bidirectional model.}
\label{fig:Image_attn}
\end{figure}


\subsubsection{Impact of Bidirectional Transformer to Visual Generation}

We conduct an ablation study to investigate the impact of transformer directionality on latent space alignment in multimodal generation tasks. As illustrated in Fig.~\ref{fig:Image_attn}, the first row shows results generated by a unidirectional Transformer, while the second row presents outputs from a BiTransformer. 
Clearly, the unidirectional Transformer variant exhibits inferior visual quality compared to the bidirectional model. 
Specifically, the performance degradation of the unidirectional Transformer becomes increasingly pronounced as the image generation prompts become more detailed—for instance, prompts in the first column contain 130 tokens, whereas those in the third column contain only 3 tokens.
This indicates that unidirectional language modeling restricts the capacity to capture cross-modal dependencies. Due to the sequential attention mechanism of LLMs, the model is inclined to overlook long-range dependencies from subsequent modalities, resulting in incomplete semantic integration and a significant reduction in generation quality. Therefore, employing a BiTransformer is essential for enhancing cross-modal dependency modeling and ensuring higher fidelity in multimodal content generation.



\section{Conclusion}

In this work, we introduced OmniBridge, a unified and modular multimodal framework designed to address two fundamental challenges in multimodal large language models.

First, to address the prohibitive computational and data costs of training unified models, OmniBridge introduces a two-stage decoupled training strategy. By leveraging powerful pre-trained LLMs with parameter-efficient adaptation, this approach circumvents the need for full-model retraining and proves to be remarkably data-efficient. Our experiments show that robust feature alignment for complex tasks like image generation, image editing, and multimodal retrieval can be achieved with a minimal number of training samples, significantly lowering the barrier for developing powerful unified models.
Second, to harmonize the often-conflicting objectives of understanding, generation, and retrieval, we introduced a novel latent space alignment mechanism powered by a Bidirectional Transformer.  Experimental results validate that this approach effectively creates a coherent, shared latent space that mitigates task interference.  This allows OmniBridge to achieve state-of-the-art or competitive performance across all three modalities within a single architecture, confirming that effective latent space alignment is a key principle for achieving genuine multi-task synergy.

In the future, OmniBridge opens several promising avenues for research, including its extension to temporal modalities like audio and video, the integration of external knowledge for more grounded generation, and the exploration of unified decoding strategies.  We believe OmniBridge represents a significant step towards building scalable, data-efficient, and truly general-purpose multimodal intelligence.


\section*{Acknowledgments}
This work was supported by the National Natural Science Foundation of China under Grant No. 62306216, the Hubei Province Technology Innovation Program under Grant No. 2024BAB043, and the Wuhan University Frontier Interdisciplinary Program under Grant No. 2042025kf0026.



\bibliographystyle{IEEEtran}
\bibliography{IEEEabrv, references}

\begin{thebibliography}{100}
\providecommand{\url}[1]{#1}
\csname url@samestyle\endcsname
\providecommand{\newblock}{\relax}
\providecommand{\bibinfo}[2]{#2}
\providecommand{\BIBentrySTDinterwordspacing}{\spaceskip=0pt\relax}
\providecommand{\BIBentryALTinterwordstretchfactor}{4}
\providecommand{\BIBentryALTinterwordspacing}{\spaceskip=\fontdimen2\font plus
\BIBentryALTinterwordstretchfactor\fontdimen3\font minus \fontdimen4\font\relax}
\providecommand{\BIBforeignlanguage}[2]{{%
\expandafter\ifx\csname l@#1\endcsname\relax
\typeout{** WARNING: IEEEtran.bst: No hyphenation pattern has been}%
\typeout{** loaded for the language `#1'. Using the pattern for}%
\typeout{** the default language instead.}%
\else
\language=\csname l@#1\endcsname
\fi
#2}}
\providecommand{\BIBdecl}{\relax}
\BIBdecl

\bibitem{llava}
H.~Liu, C.~Li, Q.~Wu \emph{et~al.}, ``Visual instruction tuning,'' \emph{Advances in neural information processing systems}, vol.~36, pp. 34\,892--34\,916, 2023.

\bibitem{sohl2015deep}
J.~Sohl-Dickstein, E.~Weiss, N.~Maheswaranathan \emph{et~al.}, ``Deep unsupervised learning using nonequilibrium thermodynamics,'' in \emph{International conference on machine learning}.\hskip 1em plus 0.5em minus 0.4em\relax pmlr, 2015, pp. 2256--2265.

\bibitem{ho2020denoising}
J.~Ho, A.~Jain, and P.~Abbeel, ``Denoising diffusion probabilistic models,'' \emph{Advances in neural information processing systems}, vol.~33, pp. 6840--6851, 2020.

\bibitem{kingma2013auto}
D.~P. Kingma, M.~Welling \emph{et~al.}, ``Auto-encoding variational bayes,'' 2013.

\bibitem{goodfellow2014generative}
I.~J. Goodfellow, J.~Pouget-Abadie, M.~Mirza \emph{et~al.}, ``Generative adversarial nets,'' \emph{Advances in neural information processing systems}, vol.~27, 2014.

\bibitem{podell2023sdxl}
D.~Podell, Z.~English, K.~Lacey \emph{et~al.}, ``Sdxl: Improving latent diffusion models for high-resolution image synthesis,'' \emph{arXiv preprint arXiv:2307.01952}, 2023.

\bibitem{esser2024scaling}
P.~Esser, S.~Kulal, A.~Blattmann \emph{et~al.}, ``Scaling rectified flow transformers for high-resolution image synthesis,'' in \emph{Forty-first international conference on machine learning}, 2024.

\bibitem{ho2022video}
J.~Ho, T.~Salimans, A.~Gritsenko \emph{et~al.}, ``Video diffusion models,'' \emph{Advances in Neural Information Processing Systems}, vol.~35, pp. 8633--8646, 2022.

\bibitem{wu2023tune}
J.~Z. Wu, Y.~Ge, X.~Wang \emph{et~al.}, ``Tune-a-video: One-shot tuning of image diffusion models for text-to-video generation,'' in \emph{Proceedings of the IEEE/CVF International Conference on Computer Vision}, 2023, pp. 7623--7633.

\bibitem{wu2021fashion}
H.~Wu, Y.~Gao, X.~Guo \emph{et~al.}, ``Fashion iq: A new dataset towards retrieving images by natural language feedback,'' in \emph{Proceedings of the IEEE/CVF Conference on computer vision and pattern recognition}, 2021, pp. 11\,307--11\,317.

\bibitem{liu2021image}
Z.~Liu, C.~Rodriguez-Opazo, D.~Teney \emph{et~al.}, ``Image retrieval on real-life images with pre-trained vision-and-language models,'' in \emph{Proceedings of the IEEE/CVF International Conference on Computer Vision}, 2021, pp. 2125--2134.

\bibitem{MagicLens}
K.~Zhang, Y.~Luan, H.~Hu \emph{et~al.}, ``Magiclens: self-supervised image retrieval with open-ended instructions,'' in \emph{Proceedings of the 41st International Conference on Machine Learning}, 2024, pp. 59\,403--59\,420.

\bibitem{chang2022webqa}
Y.~Chang, M.~Narang, H.~Suzuki \emph{et~al.}, ``Webqa: Multihop and multimodal qa,'' in \emph{Proceedings of the IEEE/CVF conference on computer vision and pattern recognition}, 2022, pp. 16\,495--16\,504.

\bibitem{DBLP:conf/iclr/LiuXL0023}
Z.~Liu, C.~Xiong, Y.~Lv \emph{et~al.}, ``Universal vision-language dense retrieval: Learning {A} unified representation space for multi-modal retrieval,'' in \emph{The Eleventh International Conference on Learning Representations, {ICLR} 2023, Kigali, Rwanda, May 1-5, 2023}.\hskip 1em plus 0.5em minus 0.4em\relax OpenReview.net, 2023.

\bibitem{luo2023end}
M.~Luo, Z.~Fang, T.~Gokhale \emph{et~al.}, ``End-to-end knowledge retrieval with multi-modal queries,'' in \emph{Proceedings of the 61st Annual Meeting of the Association for Computational Linguistics (Volume 1: Long Papers)}, 2023, pp. 8573--8589.

\bibitem{yasunaga2023retrieval}
M.~Yasunaga, A.~Aghajanyan, W.~Shi \emph{et~al.}, ``Retrieval-augmented multimodal language modeling,'' in \emph{International Conference on Machine Learning}.\hskip 1em plus 0.5em minus 0.4em\relax PMLR, 2023, pp. 39\,755--39\,769.

\bibitem{VisRAG}
S.~Yu, C.~Tang, B.~Xu \emph{et~al.}, ``Visrag: Vision-based retrieval-augmented generation on multi-modality documents,'' \emph{CoRR}, vol. abs/2410.10594, 2024.

\bibitem{NExT-GPT}
S.~Wu, H.~Fei, L.~Qu \emph{et~al.}, ``Next-gpt: Any-to-any multimodal llm,'' in \emph{International Conference on Machine Learning}.\hskip 1em plus 0.5em minus 0.4em\relax PMLR, 2024, pp. 53\,366--53\,397.

\bibitem{SEED-X}
Y.~Ge, S.~Zhao, J.~Zhu \emph{et~al.}, ``{SEED-X:} multimodal models with unified multi-granularity comprehension and generation,'' \emph{CoRR}, vol. abs/2404.14396, 2024.

\bibitem{X-VILA}
H.~Ye, D.~Huang, Y.~Lu \emph{et~al.}, ``{X-VILA:} cross-modality alignment for large language model,'' \emph{CoRR}, vol. abs/2405.19335, 2024.

\bibitem{tang2023any}
Z.~Tang, Z.~Yang, C.~Zhu \emph{et~al.}, ``Any-to-any generation via composable diffusion,'' \emph{Advances in Neural Information Processing Systems}, vol.~36, pp. 16\,083--16\,099, 2023.

\bibitem{Chameleon}
C.~Team, ``Chameleon: Mixed-modal early-fusion foundation models,'' \emph{arXiv preprint arXiv:2405.09818}, 2024.

\bibitem{Transfusion}
C.~Zhou, L.~Yu, A.~Babu \emph{et~al.}, ``Transfusion: Predict the next token and diffuse images with one multi-modal model,'' \emph{CoRR}, vol. abs/2408.11039, 2024.

\bibitem{Show-o}
J.~Xie, W.~Mao, Z.~Bai \emph{et~al.}, ``Show-o: One single transformer to unify multimodal understanding and generation,'' \emph{CoRR}, vol. abs/2408.12528, 2024.

\bibitem{Emu3}
X.~Wang, X.~Zhang, Z.~Luo \emph{et~al.}, ``Emu3: Next-token prediction is all you need,'' \emph{CoRR}, vol. abs/2409.18869, 2024.

\bibitem{li2024multimodal}
C.~Li, Z.~Gan, Z.~Yang \emph{et~al.}, ``Multimodal foundation models: From specialists to general-purpose assistants,'' \emph{Foundations and Trends{\textregistered} in Computer Graphics and Vision}, vol.~16, no. 1-2, pp. 1--214, 2024.

\bibitem{mllm2023Survey}
S.~Yin, C.~Fu, S.~Zhao \emph{et~al.}, ``A survey on multimodal large language models,'' \emph{CoRR}, vol. abs/2306.13549, 2023.

\bibitem{qwen2vl}
P.~Wang, S.~Bai, S.~Tan \emph{et~al.}, ``Qwen2-vl: Enhancing vision-language model's perception of the world at any resolution,'' \emph{CoRR}, vol. abs/2409.12191, 2024.

\bibitem{CLIP}
A.~Radford, J.~W. Kim, C.~Hallacy \emph{et~al.}, ``Learning transferable visual models from natural language supervision,'' in \emph{International conference on machine learning}.\hskip 1em plus 0.5em minus 0.4em\relax PmLR, 2021, pp. 8748--8763.

\bibitem{InstructBLIP}
W.~Dai, J.~Li, D.~Li \emph{et~al.}, ``Instructblip: Towards general-purpose vision-language models with instruction tuning,'' in \emph{Advances in Neural Information Processing Systems 36: Annual Conference on Neural Information Processing Systems 2023, NeurIPS 2023, New Orleans, LA, USA, December 10 - 16, 2023}, A.~Oh, T.~Naumann, A.~Globerson \emph{et~al.}, Eds., 2023.

\bibitem{esser2021taming}
P.~Esser, R.~Rombach, and B.~Ommer, ``Taming transformers for high-resolution image synthesis,'' in \emph{Proceedings of the IEEE/CVF conference on computer vision and pattern recognition}, 2021, pp. 12\,873--12\,883.

\bibitem{Diffusion}
P.~Sun, Y.~Jiang, S.~Chen \emph{et~al.}, ``Autoregressive model beats diffusion: Llama for scalable image generation,'' \emph{CoRR}, vol. abs/2406.06525, 2024.

\bibitem{dong2025insight}
Y.~Dong, Z.~Liu, H.-L. Sun \emph{et~al.}, ``Insight-v: Exploring long-chain visual reasoning with multimodal large language models,'' in \emph{Proceedings of the Computer Vision and Pattern Recognition Conference}, 2025, pp. 9062--9072.

\bibitem{llavaReasoner}
R.~Zhang, B.~Zhang, Y.~Li \emph{et~al.}, ``Improve vision language model chain-of-thought reasoning,'' \emph{arXiv preprint arXiv:2410.16198}, 2024.

\bibitem{transformer}
A.~Vaswani, N.~Shazeer, N.~Parmar \emph{et~al.}, ``Attention is all you need,'' \emph{Advances in neural information processing systems}, vol.~30, 2017.

\bibitem{llama}
H.~Touvron, T.~Lavril, G.~Izacard \emph{et~al.}, ``Llama: Open and efficient foundation language models,'' \emph{CoRR}, vol. abs/2302.13971, 2023.

\bibitem{VideoPoet}
D.~Kondratyuk, L.~Yu, X.~Gu \emph{et~al.}, ``Videopoet: a large language model for zero-shot video generation,'' in \emph{Proceedings of the 41st International Conference on Machine Learning}, 2024, pp. 25\,105--25\,124.

\bibitem{LLamaGen}
P.~Sun, Y.~Jiang, S.~Chen \emph{et~al.}, ``Autoregressive model beats diffusion: Llama for scalable image generation,'' \emph{arXiv preprint arXiv:2406.06525}, 2024.

\bibitem{SD}
W.~Peebles and S.~Xie, ``Scalable diffusion models with transformers,'' in \emph{Proceedings of the IEEE/CVF international conference on computer vision}, 2023, pp. 4195--4205.

\bibitem{SDXL}
D.~Podell, Z.~English, K.~Lacey \emph{et~al.}, ``{SDXL:} improving latent diffusion models for high-resolution image synthesis,'' in \emph{The Twelfth International Conference on Learning Representations, {ICLR} 2024, Vienna, Austria, May 7-11, 2024}.\hskip 1em plus 0.5em minus 0.4em\relax OpenReview.net, 2024.

\bibitem{PixArt-alpha}
J.~Chen, J.~Yu, C.~Ge \emph{et~al.}, ``Pixart-$\alpha$: Fast training of diffusion transformer for photorealistic text-to-image synthesis,'' in \emph{ICLR}, 2024.

\bibitem{DALL-E2}
A.~Ramesh, P.~Dhariwal, A.~Nichol \emph{et~al.}, ``Hierarchical text-conditional image generation with clip latents,'' \emph{arXiv preprint arXiv:2204.06125}, 2022.

\bibitem{Imagen}
C.~Saharia, W.~Chan, S.~Saxena \emph{et~al.}, ``Photorealistic text-to-image diffusion models with deep language understanding,'' \emph{Advances in neural information processing systems}, vol.~35, pp. 36\,479--36\,494, 2022.

\bibitem{ControlNet}
L.~Zhang, A.~Rao, and M.~Agrawala, ``Adding conditional control to text-to-image diffusion models,'' in \emph{Proceedings of the IEEE/CVF international conference on computer vision}, 2023, pp. 3836--3847.

\bibitem{BEIR}
N.~Thakur, N.~Reimers, A.~R{\"{u}}ckl{\'{e}} \emph{et~al.}, ``{BEIR:} {A} heterogenous benchmark for zero-shot evaluation of information retrieval models,'' \emph{CoRR}, vol. abs/2104.08663, 2021.

\bibitem{MSCOCO}
X.~Chen, H.~Fang, T.-Y. Lin \emph{et~al.}, ``Microsoft coco captions: Data collection and evaluation server,'' \emph{arXiv preprint arXiv:1504.00325}, 2015.

\bibitem{li2023blip}
J.~Li, D.~Li, S.~Savarese \emph{et~al.}, ``Blip-2: Bootstrapping language-image pre-training with frozen image encoders and large language models,'' in \emph{International conference on machine learning}.\hskip 1em plus 0.5em minus 0.4em\relax PMLR, 2023, pp. 19\,730--19\,742.

\bibitem{DreamLLM}
R.~Dong, C.~Han, Y.~Peng \emph{et~al.}, ``Dreamllm: Synergistic multimodal comprehension and creation,'' in \emph{The Twelfth International Conference on Learning Representations, {ICLR} 2024, Vienna, Austria, May 7-11, 2024}.\hskip 1em plus 0.5em minus 0.4em\relax OpenReview.net, 2024.

\bibitem{ge2023planting}
Y.~Ge, Y.~Ge, Z.~Zeng \emph{et~al.}, ``Planting a {SEED} of vision in large language model,'' \emph{CoRR}, vol. abs/2307.08041, 2023.

\bibitem{Emu}
Q.~Sun, Q.~Yu, Y.~Cui \emph{et~al.}, ``Emu: Generative pretraining in multimodality,'' in \emph{The Twelfth International Conference on Learning Representations, {ICLR} 2024, Vienna, Austria, May 7-11, 2024}.\hskip 1em plus 0.5em minus 0.4em\relax OpenReview.net, 2024.

\bibitem{Janus}
C.~Wu, X.~Chen, Z.~Wu \emph{et~al.}, ``Janus: Decoupling visual encoding for unified multimodal understanding and generation,'' \emph{CoRR}, vol. abs/2410.13848, 2024.

\bibitem{VILA-U}
Y.~Wu, Z.~Zhang, J.~Chen \emph{et~al.}, ``Vila-u: a unified foundation model integrating visual understanding and generation,'' \emph{arXiv preprint arXiv:2409.04429}, 2024.

\bibitem{sun2024autoregressive}
P.~Sun, Y.~Jiang, S.~Chen \emph{et~al.}, ``Autoregressive model beats diffusion: Llama for scalable image generation,'' \emph{CoRR}, vol. abs/2406.06525, 2024.

\bibitem{JanusFlow}
Y.~Ma, X.~Liu, X.~Chen \emph{et~al.}, ``Janusflow: Harmonizing autoregression and rectified flow for unified multimodal understanding and generation,'' \emph{CoRR}, vol. abs/2411.07975, 2024.

\bibitem{zhang2025R1_vl}
J.~Zhang, J.~Huang, H.~Yao \emph{et~al.}, ``R1-vl: Learning to reason with multimodal large language models via step-wise group relative policy optimization,'' \emph{arXiv preprint arXiv:2503.12937}, 2025.

\bibitem{mt5}
C.~Raffel, N.~Shazeer, A.~Roberts \emph{et~al.}, ``Exploring the limits of transfer learning with a unified text-to-text transformer,'' \emph{Journal of machine learning research}, vol.~21, no. 140, pp. 1--67, 2020.

\bibitem{HunyuanDiT}
Z.~Li, J.~Zhang, Q.~Lin \emph{et~al.}, ``Hunyuan-dit: A powerful multi-resolution diffusion transformer with fine-grained chinese understanding,'' \emph{arXiv preprint arXiv:2405.08748}, 2024.

\bibitem{pytorch2024}
\BIBentryALTinterwordspacing
{PyTorch-Contributors}, ``Pytorch,'' 2024, accessed: 2024. [Online]. Available: \url{https://pytorch.org}
\BIBentrySTDinterwordspacing

\bibitem{yao2024mulberry}
H.~Yao, J.~Huang, W.~Wu \emph{et~al.}, ``Mulberry: Empowering mllm with o1-like reasoning and reflection via collective monte carlo tree search,'' \emph{arXiv preprint arXiv:2412.18319}, 2024.

\bibitem{yang2025qwen3}
A.~Yang, A.~Li, B.~Yang \emph{et~al.}, ``Qwen3 technical report,'' \emph{arXiv preprint arXiv:2505.09388}, 2025.

\bibitem{zhang2023magicbrush}
K.~Zhang, L.~Mo, W.~Chen \emph{et~al.}, ``Magicbrush: A manually annotated dataset for instruction-guided image editing,'' \emph{Advances in Neural Information Processing Systems}, vol.~36, pp. 31\,428--31\,449, 2023.

\bibitem{plummer2015flickr30k}
B.~A. Plummer, L.~Wang, C.~M. Cervantes \emph{et~al.}, ``Flickr30k entities: Collecting region-to-phrase correspondences for richer image-to-sentence models,'' in \emph{Proceedings of the IEEE international conference on computer vision}, 2015, pp. 2641--2649.

\bibitem{Vlmevalkit}
H.~Duan, J.~Yang, Y.~Qiao \emph{et~al.}, ``Vlmevalkit: An open-source toolkit for evaluating large multi-modality models,'' in \emph{Proceedings of the 32nd ACM international conference on multimedia}, 2024, pp. 11\,198--11\,201.

\bibitem{MMMU}
X.~Yue, Y.~Ni, K.~Zhang \emph{et~al.}, ``Mmmu: A massive multi-discipline multimodal understanding and reasoning benchmark for expert agi,'' in \emph{Proceedings of the IEEE/CVF Conference on Computer Vision and Pattern Recognition}, 2024, pp. 9556--9567.

\bibitem{MMMUPro}
X.~Yue, T.~Zheng, Y.~Ni \emph{et~al.}, ``Mmmu-pro: A more robust multi-discipline multimodal understanding benchmark,'' \emph{arXiv preprint arXiv:2409.02813}, 2024.

\bibitem{ScienceQA}
P.~Lu, S.~Mishra, T.~Xia \emph{et~al.}, ``Learn to explain: Multimodal reasoning via thought chains for science question answering,'' \emph{Advances in Neural Information Processing Systems}, vol.~35, pp. 2507--2521, 2022.

\bibitem{MathVista}
P.~Lu, H.~Bansal, T.~Xia \emph{et~al.}, ``Mathvista: Evaluating mathematical reasoning of foundation models in visual contexts,'' in \emph{The Twelfth International Conference on Learning Representations, {ICLR} 2024, Vienna, Austria, May 7-11, 2024}.\hskip 1em plus 0.5em minus 0.4em\relax OpenReview.net, 2024.

\bibitem{MATHVision}
K.~Wang, J.~Pan, W.~Shi \emph{et~al.}, ``Measuring multimodal mathematical reasoning with math-vision dataset,'' \emph{Advances in Neural Information Processing Systems}, vol.~37, pp. 95\,095--95\,169, 2024.

\bibitem{MathVerse}
R.~Zhang, D.~Jiang, Y.~Zhang \emph{et~al.}, ``Mathverse: Does your multi-modal llm truly see the diagrams in visual math problems?'' in \emph{European Conference on Computer Vision}.\hskip 1em plus 0.5em minus 0.4em\relax Springer, 2024, pp. 169--186.

\bibitem{M3CoT}
Q.~Chen, L.~Qin, J.~Zhang \emph{et~al.}, ``M3cot: A novel benchmark for multi-domain multi-step multi-modal chain-of-thought,'' in \emph{Proceedings of the 62nd Annual Meeting of the Association for Computational Linguistics (Volume 1: Long Papers)}, 2024, pp. 8199--8221.

\bibitem{AI2D}
A.~Kembhavi, M.~Salvato, E.~Kolve \emph{et~al.}, ``A diagram is worth a dozen images,'' in \emph{Computer Vision--ECCV 2016: 14th European Conference, Amsterdam, The Netherlands, October 11--14, 2016, Proceedings, Part IV 14}.\hskip 1em plus 0.5em minus 0.4em\relax Springer, 2016, pp. 235--251.

\bibitem{ChartQA}
A.~Masry, X.~L. Do, J.~Q. Tan \emph{et~al.}, ``Chartqa: A benchmark for question answering about charts with visual and logical reasoning,'' in \emph{Findings of the Association for Computational Linguistics: ACL 2022}, 2022, pp. 2263--2279.

\bibitem{Charxiv}
Z.~Wang, M.~Xia, L.~He \emph{et~al.}, ``Charxiv: Charting gaps in realistic chart understanding in multimodal llms,'' \emph{Advances in Neural Information Processing Systems}, vol.~37, pp. 113\,569--113\,697, 2024.

\bibitem{TextVQA}
A.~Singh, V.~Natarajan, M.~Shah \emph{et~al.}, ``Towards vqa models that can read,'' in \emph{Proceedings of the IEEE/CVF conference on computer vision and pattern recognition}, 2019, pp. 8317--8326.

\bibitem{DocVQA}
C.~Clark and M.~Gardner, ``Simple and effective multi-paragraph reading comprehension,'' in \emph{Proceedings of the 56th Annual Meeting of the Association for Computational Linguistics (Volume 1: Long Papers)}, 2018, pp. 845--855.

\bibitem{InfoVQA}
M.~Mathew, V.~Bagal, R.~Tito \emph{et~al.}, ``Infographicvqa,'' in \emph{Proceedings of the IEEE/CVF Winter Conference on Applications of Computer Vision}, 2022, pp. 1697--1706.

\bibitem{OCRBench}
Y.~Liu, Z.~Li, M.~Huang \emph{et~al.}, ``Ocrbench: on the hidden mystery of ocr in large multimodal models,'' \emph{Science China Information Sciences}, vol.~67, no.~12, p. 220102, 2024.

\bibitem{SEEDBench-2-Plus}
B.~Li, Y.~Ge, Y.~Chen \emph{et~al.}, ``Seed-bench-2-plus: Benchmarking multimodal large language models with text-rich visual comprehension,'' \emph{arXiv preprint arXiv:2404.16790}, 2024.

\bibitem{BLINK}
X.~Fu, Y.~Hu, B.~Li \emph{et~al.}, ``Blink: Multimodal large language models can see but not perceive,'' in \emph{European Conference on Computer Vision}, 2024, pp. 148--166.

\bibitem{MANTIS}
D.~Jiang, X.~He, H.~Zeng \emph{et~al.}, ``{MANTIS:} interleaved multi-image instruction tuning,'' \emph{CoRR}, vol. abs/2405.01483, 2024.

\bibitem{MuirBench}
F.~Wang, X.~Fu, J.~Y. Huang \emph{et~al.}, ``Muirbench: {A} comprehensive benchmark for robust multi-image understanding,'' \emph{CoRR}, vol. abs/2406.09411, 2024.

\bibitem{MMT-Bench}
K.~Ying, F.~Meng, J.~Wang \emph{et~al.}, ``Mmt-bench: A comprehensive multimodal benchmark for evaluating large vision-language models towards multitask agi,'' in \emph{International Conference on Machine Learning}.\hskip 1em plus 0.5em minus 0.4em\relax PMLR, 2024, pp. 57\,116--57\,198.

\bibitem{RealWorldQA}
\BIBentryALTinterwordspacing
{X.AI Corp.}, ``Grok-1.5 vision preview: Connecting the digital and physical worlds with our first multimodal model,'' March 2024, accessed: 2024-03-18/19. [Online]. Available: \url{https://x.ai/blog/grok-1.5v}
\BIBentrySTDinterwordspacing

\bibitem{MME-RealWorld}
Y.~Zhang, H.~Zhang, H.~Tian \emph{et~al.}, ``Mme-realworld: Could your multimodal {LLM} challenge high-resolution real-world scenarios that are difficult for humans?'' \emph{CoRR}, vol. abs/2408.13257, 2024.

\bibitem{R-Bench}
C.~Li, J.~Zhang, Z.~Zhang \emph{et~al.}, ``R-bench: Are your large multimodal model robust to real-world corruptions?'' \emph{arXiv preprint arXiv:2410.05474}, 2024.

\bibitem{MME}
C.~Fu, P.~Chen, Y.~Shen \emph{et~al.}, ``Mme: A comprehensive evaluation benchmark for multimodal large language models,'' \emph{arXiv preprint arXiv:2306.13394}, 2023.

\bibitem{Mmbench}
Y.~Liu, H.~Duan, Y.~Zhang \emph{et~al.}, ``Mmbench: Is your multi-modal model an all-around player?'' in \emph{European Conference on Computer Vision}, 2024, pp. 216--233.

\bibitem{MMStar}
L.~Chen, J.~Li, X.~Dong \emph{et~al.}, ``Are we on the right way for evaluating large vision-language models?'' in \emph{Advances in Neural Information Processing Systems 38: Annual Conference on Neural Information Processing Systems 2024, NeurIPS 2024, Vancouver, BC, Canada, December 10 - 15, 2024}, A.~Globersons, L.~Mackey, D.~Belgrave \emph{et~al.}, Eds., 2024.

\bibitem{HallusionBench}
T.~Guan, F.~Liu, X.~Wu \emph{et~al.}, ``Hallusionbench: an advanced diagnostic suite for entangled language hallucination and visual illusion in large vision-language models,'' in \emph{Proceedings of the IEEE/CVF Conference on Computer Vision and Pattern Recognition}, 2024, pp. 14\,375--14\,385.

\bibitem{POPE}
Y.~Li, Y.~Du, K.~Zhou \emph{et~al.}, ``Evaluating object hallucination in large vision-language models,'' in \emph{Proceedings of the 2023 Conference on Empirical Methods in Natural Language Processing}, 2023, pp. 292--305.

\bibitem{Geneval}
D.~Ghosh, H.~Hajishirzi, and L.~Schmidt, ``Geneval: An object-focused framework for evaluating text-to-image alignment,'' \emph{Advances in Neural Information Processing Systems}, vol.~36, pp. 52\,132--52\,152, 2023.

\bibitem{DPG-Bench}
X.~Hu, R.~Wang, Y.~Fang \emph{et~al.}, ``Ella: Equip diffusion models with llm for enhanced semantic alignment,'' \emph{arXiv preprint arXiv:2403.05135}, 2024.

\bibitem{ye2025imgedit}
Y.~Ye, X.~He, Z.~Li \emph{et~al.}, ``Imgedit: A unified image editing dataset and benchmark,'' \emph{arXiv preprint arXiv:2505.20275}, 2025.

\bibitem{xu2024llavacot}
G.~Xu, P.~Jin, L.~Hao \emph{et~al.}, ``Llava-o1: Let vision language models reason step-by-step,'' \emph{arXiv preprint arXiv:2411.10440}, 2024.

\bibitem{deitke2024molmo_54}
M.~Deitke, C.~Clark, S.~Lee \emph{et~al.}, ``Molmo and pixmo: Open weights and open data for state-of-the-art multimodal models,'' \emph{arXiv preprint arXiv:2409.17146}, 2024.

\bibitem{wang2024measuring_245}
K.~Wang, J.~Pan, W.~Shi \emph{et~al.}, ``Measuring multimodal mathematical reasoning with math-vision dataset,'' \emph{Advances in Neural Information Processing Systems}, vol.~37, pp. 95\,095--95\,169, 2024.

\bibitem{internvl2.5}
Z.~Chen, W.~Wang, Y.~Cao \emph{et~al.}, ``Expanding performance boundaries of open-source multimodal models with model, data, and test-time scaling,'' \emph{arXiv preprint arXiv:2412.05271}, 2024.

\bibitem{Opencompass}
O.~Contributors, ``Opencompass: A universal evaluation platform for foundation models,'' 2023.

\bibitem{abdin2024phi3technicalreporthighly}
\BIBentryALTinterwordspacing
M.~Abdin, S.~Ade~Jacobs, and A.~Awan, ``Phi-3 technical report: A highly capable language model locally on your phone,'' 2024. [Online]. Available: \url{https://arxiv.org/abs/2404.14219}
\BIBentrySTDinterwordspacing

\bibitem{lu2024ovis}
S.~Lu, Y.~Li, Q.-G. Chen \emph{et~al.}, ``Ovis: Structural embedding alignment for multimodal large language model,'' \emph{arXiv preprint arXiv:2405.20797}, 2024.

\bibitem{yao2024minicpm}
Y.~Yao, T.~Yu, A.~Zhang \emph{et~al.}, ``Minicpm-v: A gpt-4v level mllm on your phone,'' \emph{arXiv preprint arXiv:2408.01800}, 2024.

\bibitem{internvl2}
Z.~Chen, W.~Wang, H.~Tian \emph{et~al.}, ``How far are we to gpt-4v? closing the gap to commercial multimodal models with open-source suites,'' \emph{Science China Information Sciences}, vol.~67, no.~12, p. 220101, 2024.

\bibitem{thawakar2025llamavo1}
\BIBentryALTinterwordspacing
O.~Thawakar, D.~Dissanayake, K.~More \emph{et~al.}, ``Llamav-o1: Rethinking step-by-step visual reasoning in llms,'' 2025. [Online]. Available: \url{https://arxiv.org/abs/2501.06186}
\BIBentrySTDinterwordspacing

\bibitem{diao2024unveiling}
H.~Diao, Y.~Cui, X.~Li \emph{et~al.}, ``Unveiling encoder-free vision-language models,'' \emph{Advances in Neural Information Processing Systems}, vol.~37, pp. 52\,545--52\,567, 2024.

\bibitem{wang2024cogvlm}
W.~Wang, Q.~Lv, W.~Yu \emph{et~al.}, ``Cogvlm: Visual expert for pretrained language models,'' \emph{Advances in Neural Information Processing Systems}, vol.~37, pp. 121\,475--121\,499, 2024.

\bibitem{mmcot}
Z.~Zhang, A.~Zhang, M.~Li \emph{et~al.}, ``Multimodal chain-of-thought reasoning in language models,'' \emph{arXiv preprint arXiv:2302.00923}, 2023.

\bibitem{shao2024visual}
H.~Shao, S.~Qian, H.~Xiao \emph{et~al.}, ``Visual cot: Advancing multi-modal language models with a comprehensive dataset and benchmark for chain-of-thought reasoning,'' \emph{Advances in Neural Information Processing Systems}, vol.~37, pp. 8612--8642, 2024.

\bibitem{deitke2025molmo}
M.~Deitke, C.~Clark, S.~Lee \emph{et~al.}, ``Molmo and pixmo: Open weights and open data for state-of-the-art vision-language models,'' in \emph{Proceedings of the Computer Vision and Pattern Recognition Conference}, 2025, pp. 91--104.

\bibitem{LDM}
R.~Rombach, A.~Blattmann, D.~Lorenz \emph{et~al.}, ``High-resolution image synthesis with latent diffusion models,'' in \emph{Proceedings of the IEEE/CVF conference on computer vision and pattern recognition}, 2022, pp. 10\,684--10\,695.

\bibitem{IF-XL}
DeepFloyd, ``Deepfloyd if,'' 2023, uRL: \url{https://huggingface.co/DeepFloyd/IF-XL-v1.0}.

\bibitem{DALL-E3}
J.~L. Betker~J, Goh~G, ``Improving image generation with better captions,'' \emph{Computer Science}, 2023.

\bibitem{LWM}
H.~Liu, W.~Yan, M.~Zaharia \emph{et~al.}, ``World model on million-length video and language with blockwise ringattention,'' \emph{CoRR}, vol. abs/2402.08268, 2024.

\bibitem{Playgroundv2.5}
D.~Li, A.~Kamko, E.~Akhgari \emph{et~al.}, ``Playground v2. 5: Three insights towards enhancing aesthetic quality in text-to-image generation,'' \emph{arXiv preprint arXiv:2402.17245}, 2024.

\bibitem{Lumina-Next}
L.~Zhuo, R.~Du, H.~Xiao \emph{et~al.}, ``Lumina-next : Making lumina-t2x stronger and faster with next-dit,'' in \emph{Advances in Neural Information Processing Systems 38: Annual Conference on Neural Information Processing Systems 2024, NeurIPS 2024, Vancouver, BC, Canada, December 10 - 15, 2024}, A.~Globersons, L.~Mackey, D.~Belgrave \emph{et~al.}, Eds., 2024.

\bibitem{Pixart-sigma}
J.~Chen, C.~Ge, E.~Xie \emph{et~al.}, ``Pixart-$\sigma$: Weak-to-strong training of diffusion transformer for 4k text-to-image generation,'' in \emph{European Conference on Computer Vision}.\hskip 1em plus 0.5em minus 0.4em\relax Springer, 2024, pp. 74--91.

\bibitem{qu2025tokenflow}
L.~Qu, H.~Zhang, Y.~Liu \emph{et~al.}, ``Tokenflow: Unified image tokenizer for multimodal understanding and generation,'' in \emph{Proceedings of the Computer Vision and Pattern Recognition Conference}, 2025, pp. 2545--2555.

\bibitem{yu2025anyedit}
Q.~Yu, W.~Chow, Z.~Yue \emph{et~al.}, ``Anyedit: Mastering unified high-quality image editing for any idea,'' in \emph{Proceedings of the Computer Vision and Pattern Recognition Conference}, 2025, pp. 26\,125--26\,135.

\bibitem{hurst2024gpt}
A.~Hurst, A.~Lerer, A.~P. Goucher \emph{et~al.}, ``Gpt-4o system card,'' \emph{arXiv preprint arXiv:2410.21276}, 2024.

\bibitem{brooks2023instructpix2pix}
T.~Brooks, A.~Holynski, and A.~A. Efros, ``Instructpix2pix: Learning to follow image editing instructions,'' in \emph{Proceedings of the IEEE/CVF conference on computer vision and pattern recognition}, 2023, pp. 18\,392--18\,402.

\bibitem{UltraEdit}
H.~Zhao, X.~S. Ma, L.~Chen \emph{et~al.}, ``Ultraedit: Instruction-based fine-grained image editing at scale,'' in \emph{NeurIPS}, 2024.

\bibitem{OmniGen}
S.~Xiao, Y.~Wang, J.~Zhou \emph{et~al.}, ``Omnigen: Unified image generation,'' in \emph{{CVPR}}.\hskip 1em plus 0.5em minus 0.4em\relax Computer Vision Foundation / {IEEE}, 2025, pp. 13\,294--13\,304.

\bibitem{Step1X-Edit}
S.~Liu, Y.~Han, P.~Xing \emph{et~al.}, ``Step1x-edit: {A} practical framework for general image editing,'' \emph{CoRR}, vol. abs/2504.17761, 2025.

\bibitem{ICEdit}
Z.~Zhang, J.~Xie, Y.~Lu \emph{et~al.}, ``In-context edit: Enabling instructional image editing with in-context generation in large scale diffusion transformer,'' \emph{CoRR}, vol. abs/2504.20690, 2025.

\bibitem{BAGEL}
C.~Deng, D.~Zhu, K.~Li \emph{et~al.}, ``Emerging properties in unified multimodal pretraining,'' \emph{CoRR}, vol. abs/2505.14683, 2025.

\bibitem{OpenCLIP}
\BIBentryALTinterwordspacing
G.~Ilharco, M.~Wortsman, R.~Wightman \emph{et~al.}, ``Openclip,'' Dec. 2021. [Online]. Available: \url{https://doi.org/10.5281/zenodo.5143773}
\BIBentrySTDinterwordspacing

\bibitem{sun2023eva}
Q.~Sun, Y.~Fang, L.~Wu \emph{et~al.}, ``Eva-clip: Improved training techniques for clip at scale,'' \emph{arXiv preprint arXiv:2303.15389}, 2023.

\bibitem{chen2024internvl}
Z.~Chen, J.~Wu, W.~Wang \emph{et~al.}, ``Internvl: Scaling up vision foundation models and aligning for generic visual-linguistic tasks,'' in \emph{Proceedings of the IEEE/CVF conference on computer vision and pattern recognition}, 2024, pp. 24\,185--24\,198.

\bibitem{multi2024m3}
M.-L. M.-F. Multi-Granularity, ``M3-embedding: Multi-linguality, multi-functionality, multi-granularity text embeddings through self-knowledge distillation,'' 2024.

\bibitem{zhou2024megapairs}
J.~Zhou, Z.~Liu, Z.~Liu \emph{et~al.}, ``Megapairs: Massive data synthesis for universal multimodal retrieval,'' \emph{arXiv preprint arXiv:2412.14475}, 2024.

\end{thebibliography}


\vfill

\end{document}